\documentclass[10pt,twocolumn,letterpaper]{article}

\usepackage{cvpr}              
\usepackage{multirow}
\usepackage[normalem]{ulem}
\useunder{\uline}{\ul}{}
\usepackage{booktabs}

\usepackage{color}
\usepackage{xcolor}
\usepackage{graphicx}
\usepackage{colortbl}
\usepackage{amsmath,amssymb} 
\usepackage{makecell}
\usepackage{bbding}
\usepackage{tabularx}
\usepackage{bm}

\usepackage[utf8]{inputenc} 
\usepackage[T1]{fontenc}    
\usepackage{url}            
\usepackage{amsfonts}       
\usepackage{nicefrac}       
\usepackage{microtype}      
\usepackage{amsthm}
\usepackage{pifont}   
\usepackage[accsupp]{axessibility}  

\usepackage{graphicx}
\allowdisplaybreaks[4]
\usepackage{wrapfig}
\usepackage{tikz}
\newcommand*{\circled}[1]{\lower.7ex\hbox{\tikz\draw (0pt, 0pt)%
    circle (.36em) node {\makebox[1em][c]{\small #1}};}}

\usepackage{listings}

\definecolor{dkgreen}{rgb}{0,0.6,0}
\definecolor{gray}{rgb}{0.5,0.5,0.5}
\definecolor{mauve}{rgb}{0.58,0,0.82}

\definecolor{cvprblue}{rgb}{0.21,0.49,0.74}
\usepackage[pagebackref,breaklinks,colorlinks,allcolors=cvprblue]{hyperref}

\def\paperID{} 
\def\confName{CVPR}
\def\confYear{2026}

\title{Cross-Scale Pansharpening via ScaleFormer and the PanScale Benchmark}

\author{
Ke Cao\textsuperscript{1,2,*}, Xuanhua He\textsuperscript{3,*,\dag}, Xueheng Li\textsuperscript{1,2}, Lingting Zhu\textsuperscript{4}, Yingying Wang\textsuperscript{5}, \\ Ao Ma\textsuperscript{6}, Zhanjie Zhang\textsuperscript{7}, Man Zhou\textsuperscript{2,\ddag}, Chengjun Xie\textsuperscript{1,2}, Jie Zhang\textsuperscript{1,2,\ddag} \\
\textsuperscript{1}HFIPS, Chinese Academy of Sciences,\textsuperscript{2}University of Science and Technology of China, \\
\textsuperscript{3}The Hong Kong University of Science and Technology, \textsuperscript{4}The University of Hong Kong, \\
\textsuperscript{5}Xiamen University, \textsuperscript{6}JD.com, \textsuperscript{7}Zhejiang University \\
{\tt\small \{caoke200820,hexuanhua,manman\}@mail.ustc.edu.cn, zhangjie@iim.ac.cn}
}

\begin{document}
\maketitle

\begingroup
\renewcommand\thefootnote{}%
\footnotetext{%
\textsuperscript{*} Equal contribution.\par
\hspace{\parindent}\textsuperscript{\dag} Project leader.\par
\hspace{\parindent}\textsuperscript{\ddag} Corresponding author.%
}
\endgroup

\begin{abstract}
Pansharpening aims to generate high-resolution multi-spectral images by fusing the spatial detail of panchromatic images with the spectral richness of low-resolution MS data. However, most existing methods are evaluated under limited, low-resolution settings, limiting their generalization to real-world, high-resolution scenarios. To bridge this gap, we systematically investigate the data, algorithmic, and computational challenges of cross-scale pansharpening. We first introduce PanScale, the first large-scale, cross-scale pansharpening dataset, accompanied by PanScale-Bench, a comprehensive benchmark for evaluating generalization across varying resolutions and scales. To realize scale generalization, we propose ScaleFormer, a novel architecture designed for multi-scale pansharpening. ScaleFormer reframes generalization across image resolutions as generalization across sequence lengths: it tokenizes images into patch sequences of the same resolution but variable length proportional to image scale. A Scale-Aware Patchify module enables training for such variations from fixed-size crops. ScaleFormer then decouples intra-patch spatial feature learning from inter-patch sequential dependency modeling, incorporating Rotary Positional Encoding to enhance extrapolation to unseen scales. Extensive experiments show that our approach outperforms SOTA methods in fusion quality and cross-scale generalization. The datasets and source code are available at \href{https://github.com/caoke-963/ScaleFormer}{https://github.com/caoke-963/ScaleFormer}.
\end{abstract}    
\section{Introduction}
\label{sec:intro}
\begin{figure}[htpb]
    \centering
    \includegraphics[width=\linewidth]{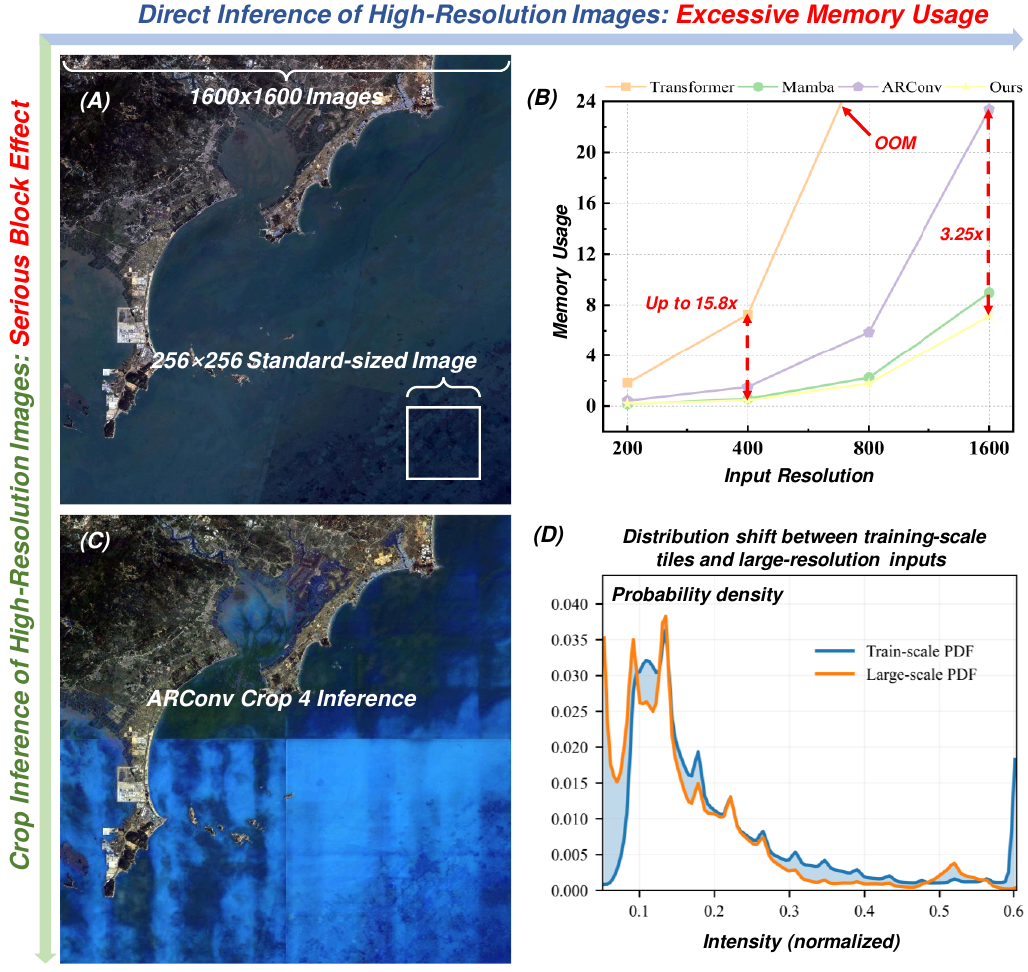}
    \caption{\label{intro4_1030}Challenges in cross-scale pansharpening. (a) Learning/modeling at limited resolution while supporting inference on high-resolution images. (b) GPU memory increases rapidly with resolution, especially for Transformer-based models. (c) Tiled inference can introduce block artifacts in some methods. (d) A scale-induced distribution shift exists between training-scale inputs and large-scale inference.}
\end{figure}
High-resolution multi-spectral (HRMS) imagery is in high demand across various fields such as environmental monitoring and precision agriculture, due to its ability to provide fine spatial detail and rich spectral information. However, current remote sensing technologies face fundamental limitations of satellite sensor hardware, making it difficult to acquire HRMS images directly. As a result, existing systems typically capture high-resolution panchromatic (PAN) images that contain fine spatial detail, alongside low-resolution multi-spectral (LRMS) images that preserve spectral characteristics. To overcome this constraint, pansharpening has been developed to utilize PAN and LRMS images to generate a fusion that combines the advantages of both. Essentially, pansharpening can be viewed as a guided super-resolution process, where the fine-grained spatial details from the PAN are leveraged to enhance the LRMS while preserving its spectral fidelity. Despite recent advances in CNN- and Transformer-based pansharpening~\cite{zhou2022sfiin,zhou2022innformer,tan2024hfin_cvpr24,zhang2024pan,wang2025arconv_cvpr25} have shown promising progress, they still face several challenges, particularly in handling images with high spatial resolutions and across varying scales. These limitations hinder their generalization and robustness in real-world applications involving multi-scale or high-resolution data.

In practical remote-sensing deployments, inference is typically performed at resolutions far exceeding the training crop size, which exposes two fundamental bottlenecks in existing approaches. \textbf {First, computation and memory scale steeply with input size}: when moving from common training crops (200-256 pixels) to 800, 1600, or even 2000 pixels kernel coverage, feature-map dimensions, and attention sequence length increase in lockstep, leading to a sharp rise in computational cost. Engineers are then forced to adopt tiled inference, which introduces boundary discontinuities and visible blocking artifacts. As illustrated in Fig.~\ref{intro4_1030}(a) and (b), panel (a) contrasts a typical training crop with our 1600-pixel inference image, while panel (b) shows the sharp rise in memory consumption with increasing resolution, particularly for Transformer-based models, which often exceed consumer-grade GPU capacity already at 800-pixel inputs. Fig.~\ref{intro4_1030}(c) highlights severe block artifacts observed with ARConv~\cite{wang2025arconv_cvpr25} under tiled inference, leading to a marked drop in fusion quality. \textbf {Second, cross-scale generalization is weak}: because most models are trained on a single, low-resolution regime, upscaling at test time induces pronounced shifts in input statistics (means, variances, and spectral composition) relative to the training distribution, degrading feature alignment and spectral fidelity and thus causing systematic performance decline~\cite{chu2022improving}. To quantify this scale-induced shift, we compare pixel PDFs between training-scale tiles and large-scale inference. As resolution increases from 200 to 1600, the luminance distribution exhibits systematic shifts in shadows and highlights, which substantially impact fusion quality. \textbf {Compounding these issues is the lack of standardized datasets and evaluation protocols targeting multi-scale and high-resolution pansharpening}: existing resources~\cite{deng2022pancollection, meng2020NBU, vivone2021PAirMax} provide limited scale diversity and spatial resolution, falling short of real-world needs and constraining the development and fair assessment of advanced methods under high-resolution, cross-scale conditions.
\begin{figure}[ht]
    \centering
    \includegraphics[width=\linewidth]{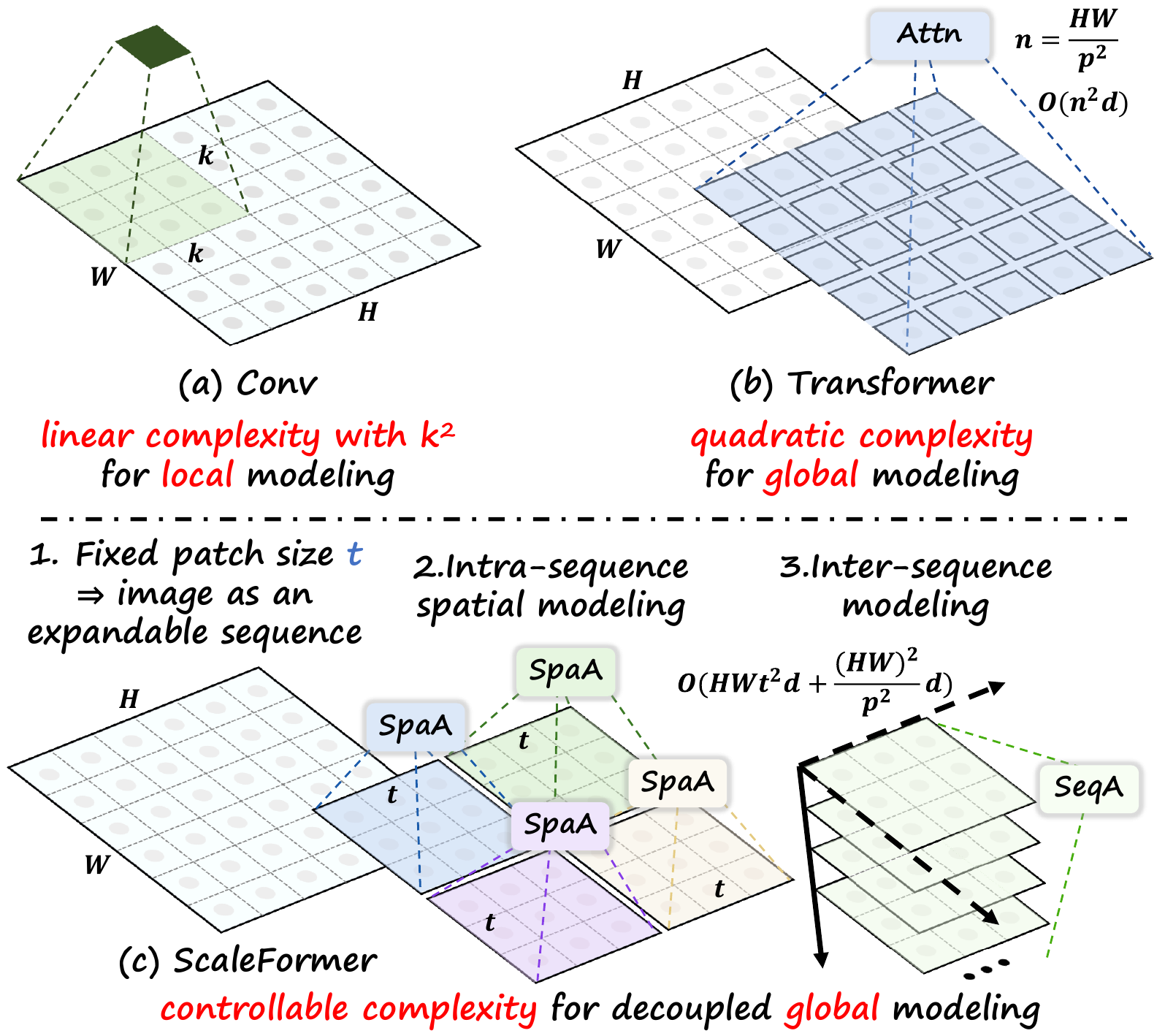}
    \caption{\label{intro_3methods}Comparison of convolution, Transformer, and the proposed ScaleFormer. (a) Convolution: linear in image size but with a limited receptive field on high-resolution inputs; enlarging kernels helps but adds quadratic cost in kernel size. (b) Transformer: uniform patchification enables global context, yet self-attention scales quadratically with patch count, driving computation up on large images. (c) ScaleFormer (ours): introduces a sequence axis, factorizing global modeling into fixed-size spatial processing plus a scalable sequence dimension, yielding controllable complexity and mitigating scale-induced distribution shifts for robust performance across resolutions.}
\end{figure}

To address these challenges, we introduce PanScale-Datasets, the first cross-scale remote sensing image fusion dataset specifically designed for pansharpening in multi-resolution and high-resolution scenarios. The dataset comprises satellite imagery from three types of remote sensing platforms, with native spatial resolutions ranging from 0.5 meters to 15 meters, thereby covering a broad spectrum of real-world sensing conditions. To simulate practical multi-scale processing demands, we construct a series of reduced-resolution and full-resolution test sets, ranging in size from $200 \times 200$ to $2000 \times 2000$ pixels, enabling rigorous evaluation under diverse input resolutions and spatial scales. In conjunction with the dataset, we propose a comprehensive evaluation suite, PanScale-Bench, which supports standardized and fair benchmarking across different methods. It integrates a variety of reference-based and no-reference metrics to comprehensively assess fusion quality at varying scales, offering a unified platform for performance comparison and analysis within the community.

Building upon this, we introduce ScaleFormer, a framework designed for robust remote-sensing image fusion across diverse scales and resolutions. As outlined in Fig.~\ref{intro_3methods}, the core idea is to reinterpret resolution changes as sequence-length changes: local patches at a fixed spatial size serve as statistically stable tokens, and only the sequence length grows linearly with image size. Concretely, we propose Scale-Aware Patchify (SAP), which employs bucketed window sampling during training to expose the model to multiple effective sequence lengths, thereby stabilizing per-token mean and variance and narrowing the training-inference distribution gap. We further adopt decoupled intra-spatial and inter-sequence modeling, and incorporate RoPE~\cite{su2024rope} to enhance extrapolation to unseen sequence lengths corresponding to larger resolutions. This reformulation markedly reduces VRAM and GFLOPs at large scales and avoids the tiling artifacts common in blockwise inference. Extensive experiments on the PanScale dataset demonstrate that ScaleFormer achieves accurate and efficient fusion over a wide range of scales and resolutions, confirming its effectiveness and advantage for multi-scale pansharpening.

Our contributions can be summarized as follows:
\begin{itemize}
\item We present PanScale, the first cross-scale pansharpening dataset, together with PanScale-Bench, a comprehensive evaluation benchmark. This standardized platform enables systematic testing and assessment of pansharpening methods in diverse multi-scale and high-resolution scenarios encountered in real-world applications.

\item We propose ScaleFormer, a framework designed for robust multi-scale pansharpening. It reframes the problem of generalizing to unseen image resolutions as generalizing to unseen sequence lengths. ScaleFormer achieves this by decoupling the modeling of fine-grained spatial structures within image patches and the scale-adaptive sequential dependencies between these patches, leading to stable performance across various input resolutions.

\item We incorporate bucket training strategy and rotary positional encoding to promote better generalization to unseen scales. These techniques encourage the model to adapt to varying resolutions beyond those seen during training.

\item Extensive experiments across multiple datasets within the PanScale demonstrate that our method achieves accurate and efficient fusion at multiple scales and maintains high performance even when processing ultra-high-resolution inputs, highlighting its strong generalization ability and practical applicability in real-world multi-scale fusion.

\end{itemize}
\section{Related Work}
\label{sec:relatedwork}
Please refer to the Appendix~\ref{Appendix Related Work}.
\section{Proposed Datasets and Benchmark}
\label{sec:Datasets}
\subsection{Data Collection and Preprocessing}
To construct the PanScale Datasets, we leveraged the capabilities of Google Earth Engine (GEE)~\cite{gorelick2017google} to systematically acquire and preprocess multi-source remote sensing data. The overall data preparation pipeline and detailed process are in the Appendix~\ref{Detailed Information for Preprocessing}.

\subsection{Dataset composition and benchmark design}
Most existing pansharpening methods are evaluated on relatively small-scale datasets with fixed resolutions, which limits their applicability to real-world scenarios involving high-resolution satellite imagery. To comprehensively assess the generalization ability of fusion methods on multi-scale, high-resolution inputs, we designed the PanScale dataset, which consists of three sub-datasets. Each sub-dataset includes a training set, multi-scale reduced-resolution test sets, and multi-scale full-resolution test sets. The detailed configurations of these datasets are summarized in Fig.~\ref{dataset_ratio}. A more comprehensive description can be found in Appendix~\ref{Detailed Information for Dataset Construction}. Furthermore, the benchmark design is provided in Appendix~\ref{Appendix Benchmark Design}.
\begin{figure*}[ht]
    \centering
    \includegraphics[width=0.95\textwidth]{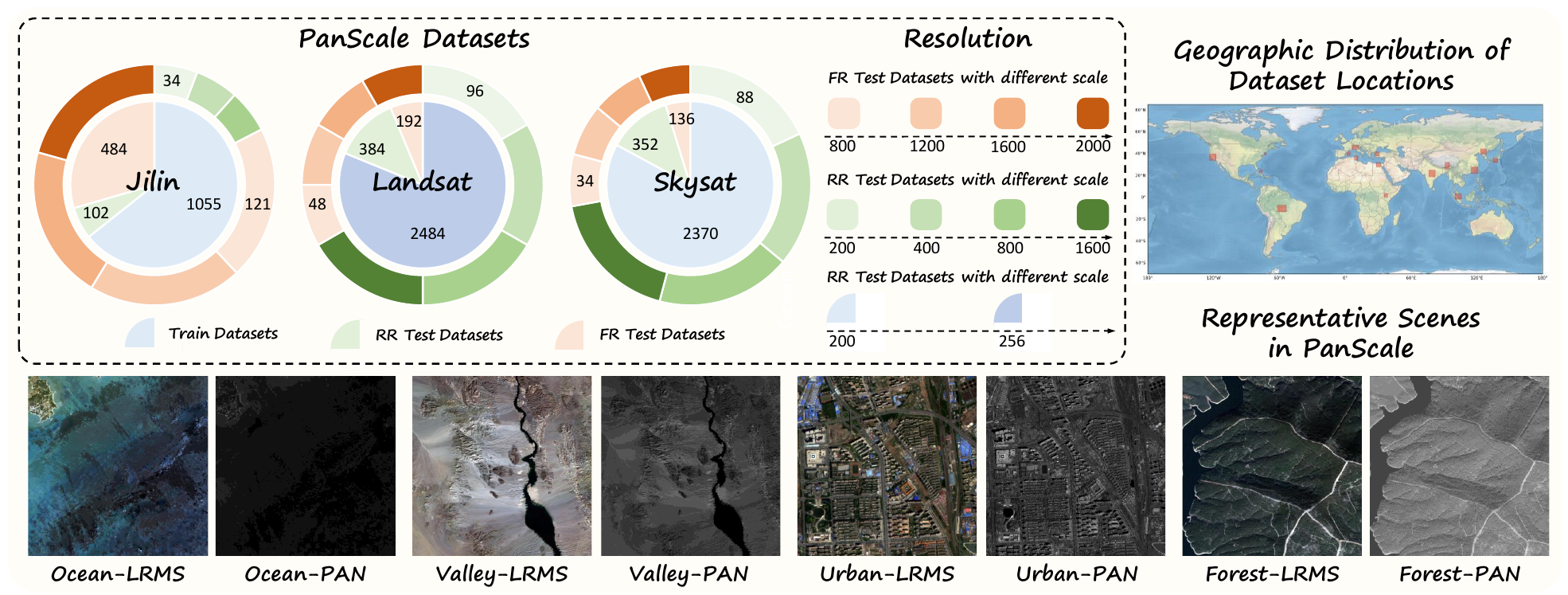}
    \caption{\label{dataset_ratio}PanScale dataset: composition and distribution, resolution range, geographic sampling locations, and representative scene visualizations.}
\end{figure*}

Table~\ref{table_dataset_compare} presents a comparison between the PanScale dataset and several widely used pansharpening datasets, focusing on the following aspects: data type, whether the data is real, the inclusion of a training set, and support for Multi-Resolution Testing (MRT) and Large Scale Testing (LST). It is evident that our dataset is unique in its simultaneous emphasis on both training data and large-scale testing, and is the only one that includes cross-scale evaluation.
\begin{table}[!htb]
    \centering
    \small
    \renewcommand{\arraystretch}{1.2}
    \renewcommand{\tabcolsep}{3pt}
    \caption{Comparison between the PanScale dataset and widely used pansharpening datasets, focusing on data type, whether the data is real, the inclusion of a training set, and support for Multi-Resolution Testing (MRT) and Large Scale Testing (LST).}
    \label{table_dataset_compare}
    \begin{tabular}{ccccccccc}
    \hline
    Datasets & Type & Real data & Train & MRT & LST\\ \hline
    PanCollection\cite{deng2022pancollection} & RR and FR & \ding{51} & \ding{51} & \ding{55} & \ding{55}\\ 
    NBU\cite{meng2020NBU}           & FR        & \ding{51} & \ding{51} & \ding{55} & \ding{55}\\ 
    PAirMax\cite{vivone2021PAirMax}       & RR and FR & \ding{51} & \ding{55} & \ding{55} & \ding{51}\\ 
    \rowcolor{gray!20}
    PanScale      & RR and FR & \ding{51} & \ding{51} & \ding{51} & \ding{51}\\ 
    \hline
    \end{tabular}
\end{table}

\section{Proposed Methods}
\label{sec:Methods}
\begin{figure*}[h]
    \centering
    \includegraphics[width=0.9\textwidth]{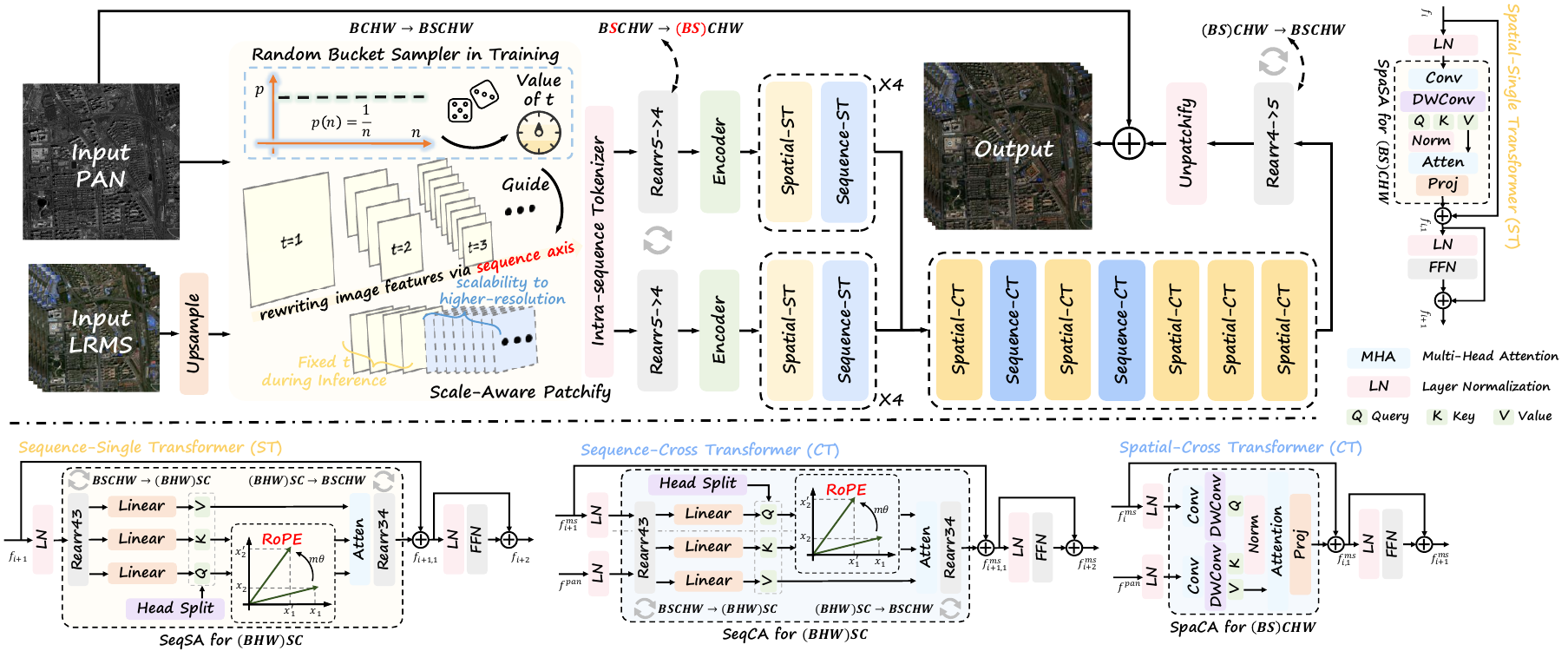}
    \caption{\label{methods_main}The overall architecture of ScaleFormer, which primarily consists of three key components: the Scale-Aware Patchify module implementing the bucket sampling strategy, the Single Transformer Modules for intra-modal feature modeling, and the Cross Transformer Modules for inter-modal feature interaction and fusion.}
\end{figure*}
\subsection{Overview}
Fig.~\ref{methods_main} illustrates the overall architecture of the proposed ScaleFormer, which is composed of three primary components: (1) the Scale-Aware Patchify module that incorporates a bucket-based training strategy to promote scale diversity and robustness; (2) Single-Transformer modules consisting of the Spatial-Transformer and Sequence-Transformer, each dedicated to modeling spatial and sequential dependencies independently; and (3) Cross-Transformer modules that include the Spatial-Cross Transformer and Sequence-Cross Transformer, which facilitate cross-modal interactions through cross-attention mechanisms along spatial and sequential dimensions, respectively. Given an input panchromatic image $\mathbf {P}\in {{\mathbb{R}}^{H\times W\times 1}}$ and a corresponding low-resolution multi-spectral (LRMS) image ${\mathbf{L}_{in}}\in {{\mathbb{R}}^{(H/r)\times (W/r)\times C}}$, where $r$ denotes the upsampling ratio between the two modalities, we first apply bicubic interpolation to $\mathbf{L}_{in}$ to obtain an upsampled multi-spectral image $\mathbf{L}\in {{\mathbb{R}}^{H\times W\times C}}$ that matches the spatial dimensions of the PAN image. 
The Scale-Aware Patchify module performs randomized sliding window sampling at various patch sizes to enable multi-scale representation learning. The extracted patches are then concatenated along a newly introduced sequence dimension, effectively converting spatially aligned inputs into a sequence of multi-scale tokens suitable for subsequent transformer-based modeling:
\begin{equation}\label{}
{\mathbf{P}_{5d}},{\mathbf{L}_{5d}}=SAP(\mathbf{P}),SAP(\mathbf{L})
\end{equation}
where $SAP$ represents the operation of Scale-Aware Patchify, $\mathbf{{P}_{5d}}\in {{\mathbb{R}}^{B\times T\times C\times h\times w}}$,${\mathbf{L}_{5d}}\in {{\mathbb{R}}^{B\times T\times C\times h\times w}}$,$h=H/T$,$w=W/T$. To facilitate subsequent processing, we reshape ${\mathbf{P}_{5d}}$ and ${\mathbf{L}_{5d}}$ by merging the batch and sequence dimensions. These are then passed through a lightweight feature encoder to extract deep feature representations for each modality, denoted as $\mathbf{f}_{4d}^{pan}$ and $\mathbf{f}_{4d}^{ms}$:
\begin{equation}\label{}
\mathbf{f}_{4d}^{pan},\mathbf{f}_{4d}^{ms}=Enc({{R}_{54}}({\mathbf{P}_{5d}})),Enc({{R}_{54}}({\mathbf{L}_{5d}}))
\end{equation}
Here, ${{R}_{54}}$ denotes the tensor rearrangement operation that transforms a 5D representation into a 4D format. Following this, we employ a cascaded structure consisting of a Spatial Transformer and a Sequence Transformer to perform hierarchical feature extraction across both the spatial and the newly introduced scale dimensions within each modality.

To enhance cross-modal interaction and facilitate more effective feature fusion, we further introduce Spatial-Cross Transformer and Sequence-Cross Transformer modules. These modules enable bidirectional information exchange between the panchromatic and multi-spectral features along spatial and scale dimensions, respectively, yielding a fused representation denoted as $\mathbf{F}_{4d}^{fus}$. To accurately reconstruct the high-resolution multi-spectral output, we apply an inverse rearrangement operation to decompose the sequence dimension of $\mathbf{F}_{4d}^{fus}$, followed by a patch reassembly process. This final step produces the desired high-resolution multi-spectral image, denoted as ${\mathbf{H}_{out}}$.

\subsection{Scale-Aware Patchify}
To mitigate the distribution shift that arises as input resolution increases, we introduce a Scale-Aware Patchify (SAP) module that equips the model with explicit awareness of spatial scale. SAP rewrites local image evidence into an additional sequence axis, enabling subsequent components to perform spatial modeling within patches while conducting inter-sequence modeling along the sequence. This decoupling preserves spatial granularity inside each window and delegates scale variation to the sequence length, improving robustness when the input size changes. Inspired by recent progress in multi-resolution training for video generation~\cite{podell2023sdxl,chen2023pixarta,zheng2024open}, we further adopt bucketed sampling to expose the network to unseen sequence lengths during training. Concretely, given co-registered panchromatic (PAN) and multispectral (MS) inputs, we randomly sample a bucket index
$t$ that determines a pre-defined window size $w(t)$. A Patch-to-Sequence Tokenizer then partitions the inputs using $w(t)$, producing a token sequence whose length encodes the effective scale. This procedure increases the model’s coverage of resolution variability, narrowing the gap between the training distribution and the test-time operating points. In inference, the system deterministically uses a fixed window size (probability 100\%), so that higher spatial resolutions are handled simply by lengthening the sequence along the scale axis while keeping per-token statistics stable. In effect, SAP prevents the mean and variance drift that typically accompanies scale changes and allows explicit and independent modeling of spatial and scale correlations. This design yields strong generalization across diverse and previously unseen input resolutions, aligning with the goals of cross-scale pansharpening in ScaleFormer and the evaluation protocols of the PanScale Benchmark.
\subsection{Single Transformer Modules}
Following the Scale-Aware Patchify, the input images are transformed into fixed-size spatial patches while introducing an additional sequence dimension that encodes scale-aware structural information. To facilitate effective learning across spatial and sequential representations, we introduce a Single-Transformer module consisting of a Spatial Transformer and a Sequence Transformer. These modules are specialized to independently capture dependencies within the spatial and sequential domains, respectively. The Spatial Transformer focuses on modeling local and global spatial relationships within each patch, enabling the network to extract rich spatial features. In parallel, the Sequence Transformer operates along the newly introduced sequence dimension, learning cross-patch correlations and capturing scale-aware contextual dependencies. This separation of concerns allows the model to decouple spatial encoding from scale modeling, thus improving both representation capacity and scalability across varying input resolutions. Specifically, given an input feature map ${\mathbf{f}_{i,1}}$, the Single-Transformer module processes it sequentially along the spatial and sequence dimensions. This procedure can be formally described as follows:
\begin{align}
  & {\mathbf{f}_{i,1}}={\mathbf{f}_{i}}+S{{A}_{spa}}(LN({\mathbf{f}_{i}})) \\ 
 & {\mathbf{f}_{i+1}}={\mathbf{f}_{i,1}}+FFN(LN({\mathbf{f}_{i,1}})) \\ 
 & {\mathbf{f}_{i+1,1}}={\mathbf{f}_{i+1}}+S{{A}_{seq}}(LN({\mathbf{f}_{i+1}})) \\ 
 & {\mathbf{f}_{i+2}}={\mathbf{f}_{i+1,1}}+FFN(LN({\mathbf{f}_{i+1,1}})) 
\end{align}
where $S{{A}_{spa}}$ denotes the self-attention operation applied along the spatial dimensions, and $LN$ represents the Layer Normalization. Subsequently, $S{{A}_{seq}}$ applies self-attention along the sequence dimension. To facilitate this, we perform a rearrangement operation that merges the batch and spatial dimensions into a single axis. Additionally, we incorporate Rotary Position Embedding (RoPE)~\cite{su2024rope} into the mapped queries and keys, which significantly enhances the model’s ability for scale extrapolation by encoding continuous relative positional information.
\subsection{Cross Transformer Modules}
Similarly, the Cross Transformer module comprises a Spatial Cross Transformer and a Sequence Cross Transformer. Specifically, given the input feature maps $\mathbf{f}_{i}^{ms}$ and ${\mathbf{f}^{pan}}$, the stacked Spatial-Cross Transformer and Sequence-Cross Transformer perform a series of cross-attention operations to facilitate effective interaction and information fusion between the multi-spectral and panchromatic features. The computation can be expressed as:
\begin{align}
  & \mathbf{f}_{i,1}^{ms}=\mathbf{f}_{i}^{ms}+C{{A}_{spa}}(LN(\mathbf{f}_{i}^{ms}),LN({\mathbf{f}^{pan}})) \\ 
 & \mathbf{f}_{i+1}^{ms}=\mathbf{f}_{i,1}^{ms}+FFN(LN(\mathbf{f}_{i}^{ms})) \\ 
 & \mathbf{f}_{i+1,1}^{ms}=\mathbf{f}_{i+1}^{ms}+C{{A}_{seq}}(LN(\mathbf{f}_{i+1}^{ms}),LN({\mathbf{f}^{pan}})) \\ 
 & \mathbf{f}_{i+2,1}^{ms}=\mathbf{f}_{i+1,1}^{ms}+FFN(LN(\mathbf{f}_{i+1,1}^{ms}))  
\end{align}
Here, $LN$ denotes Layer Normalization. $C{{A}_{spa}}$ refers to the cross-attention mechanism along the spatial dimensions, while $C{{A}_{seq}}$ operates on the sequence dimension. Similar to the single-modality attention, we apply a rearrangement operation during $C{{A}_{seq}}$ to merge the batch and spatial dimensions into a single axis. RoPE is also injected into the projected queries and keys, which enhances the model’s capacity for generalizing across scales by encoding relative positional information.

\begin{table*}[h]
\centering
\normalsize
\renewcommand{\arraystretch}{1.1}
\renewcommand{\tabcolsep}{3pt}
\caption{Quantitative Results on PanScale Datasets. The best results are highlighted in \textbf{bold}. The second-best results are highlighted in \underline{underline}. $\uparrow$ indicates that the larger the value, the better the performance, and $\downarrow$ indicates that the smaller the value, the better the performance. ARConv uses tiled inference due to OOM, and other methods are evaluated with train–inference scale misalignment.\label{table_main}}
\resizebox{\linewidth}{!}
{
\begin{tabular}{c|cccc|cccc|cccc}
\hline
& \multicolumn{4}{c|}{Jilin-Ave}                                        & \multicolumn{4}{c|}{Landsat-Ave}                                      & \multicolumn{4}{c}{Skysat-Ave}    
\\ \cline{2-13} 
\multirow{-2}{*}{Method} 
& PSNR$\uparrow$      & SSIM$\uparrow$     & ERGAS$\downarrow$        & Q $\uparrow$ 
& PSNR$\uparrow$      & SSIM$\uparrow$     & ERGAS$\downarrow$        & Q $\uparrow$ 
& PSNR$\uparrow$      & SSIM$\uparrow$    & ERGAS$\downarrow$         & Q $\uparrow$  \\ \hline
GS & 25.4386 & 0.7055 & 5.6091 & 0.5849 & 35.0781 & 0.8952 & 4.5954 & 0.5873 & 39.0558 & 0.9317 & 2.3673 & 0.6371 \\
IHS & 25.5950 & 0.7023 & 5.8071 & 0.5803 & 35.8777 & 0.8915 & 4.3234 & 0.5554 & 39.5900 & 0.9304 & 2.1756 & 0.6271  \\
GFPCA & 25.6134 & 0.7185 & 5.0460 & 0.4509 & 36.2597 & 0.8963 & 3.6041 & 0.4747 & 37.6663 & 0.9280 & 2.4973 & 0.4773 \\ \hline
MSDCNN & 36.5711 & 0.9574 & 1.4965 & 0.9092 & 38.8173 & 0.9592 & 2.6570 & 0.6737 & 42.5298 & 0.9696 & 2.5983 & 0.7261 \\
SFINet & 36.2845 & 0.9523 & 1.6129 & 0.8958 & 38.8276 & 0.9567 & 2.8075 & 0.6739 & 41.7457 & 0.9518 & 2.9933 & 0.6950 \\
MSDDN & 36.9407 & 0.9646 & 1.4326 & 0.9195 & 37.6899 & 0.9346 & 2.9023 & 0.6421 & 41.0392 & 0.9456 & 2.8546 & 0.6949 \\
PanFlowNet & 36.3326 & 0.9566 & 1.5554 & 0.9067 & 38.2158 & 0.9433 & 3.2700 & 0.6669 & 41.0219 & 0.9530 & 2.9228 & 0.6728 \\
HFIN 
& 38.0009 & \underline{0.9698} & 1.2754 & \underline{0.9284} 
& \underline{40.2110} & \underline{0.9666} & \underline{2.3230} & \underline{0.7216} 
& \underline{43.9592} & 0.9658 & 2.2738 & \underline{0.7707} \\
Pan-mamba & 35.5514 & 0.9480 & 1.6963 & 0.8966 & 36.7272 & 0.9206 & 3.2483 & 0.6437 & 41.3851 & 0.9493 & 2.7667 & 0.7129 \\
ARConv
& \underline{38.2340} & 0.9697 & \underline{1.2462} & 0.9291 
& 39.6646 & 0.9638 & 2.4441 & 0.6866 
& 43.4037 & \underline{0.9797} & \underline{2.0629} & 0.7564 \\
\rowcolor{gray!20}
Ours & \textbf{39.2932} & \textbf{0.9761} & \textbf{1.0991} & \textbf{0.9385} & \textbf{41.0360} & \textbf{0.9711} & \textbf{2.1354} & \textbf{0.7395} & \textbf{44.6546} & \textbf{0.9827} & \textbf{1.7293} & \textbf{0.7835} \\
\hline
\end{tabular}
}
\end{table*}
\begin{figure*}[h]
    \centering
    \includegraphics[width=0.90\textwidth]{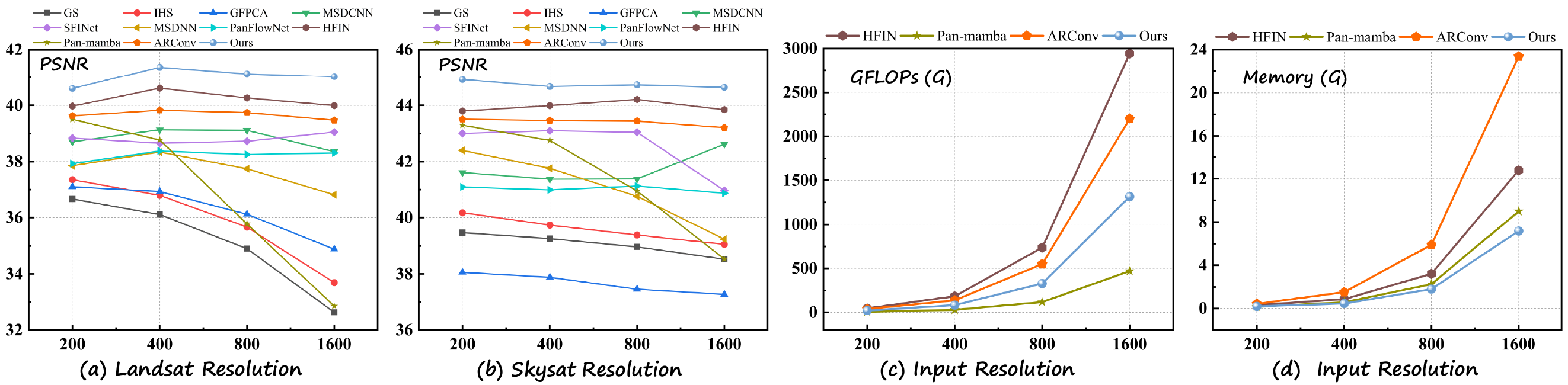}
    \caption{\label{fig_scale}Comparison results across increasing resolution scales. The four subplots respectively depict: (a) PSNR score on the Landsat dataset, (b) PSNR score on the Skysat dataset, (c) model computational complexity (measured in GFLOPs) as a function of input resolution, and (d) Memory variation concerning input resolution.}
\end{figure*}

\subsection{Loss Function}
Following common practices in the field~\cite{li2025pan,li2025exploring,tan2024hfin_cvpr24,cao2024shuffle}, we employ the L1 norm as the loss function during training, which measures the pixel-wise discrepancy between the output image ${\mathbf{H}_{out}}$ and the ground truth $\mathbf{G}$. The loss can be formally defined as:
\begin{equation}\label{}
\mathbf{L} = {{\left\| {\mathbf{H}_{out}}-\mathbf{G} \right\|}_{1}}
\end{equation}

\section{Experiments}
\label{sec:experiments}
\subsection{Implementation Details}
\label{Implementation Details}
In the comparative experiments, we selected a range of representative traditional methods as well as state-of-the-art deep learning techniques, including GS~\cite{GS}, IHS~\cite{IHS}, GFPCA~\cite{GFPCA}, MSDCNN~\cite{msdcnn}, SFINet~\cite{zhou2022sfiin}, MSDDN~\cite{he2023msddn}, PanFlowNet~\cite{yang2023panflownet}, HFIN~\cite{tan2024hfin_cvpr24}, Pan-mamba~\cite{he2025panmamba}, and ARConv~\cite{wang2025arconv_cvpr25}. These methods encompass a wide spectrum of approaches, ranging from classical image fusion techniques to the most recent deep learning-based models, thereby providing a comprehensive comparison for evaluating the performance of our proposed method. To ensure consistency and fairness, all methods are trained separately on the three training sets of the PanScale-Datasets and assessed on the corresponding RR and FR multi-resolution test sets. Additionally, the evaluation metrics are derived from the PanScale-Bench, providing a comprehensive and standardized basis for performance assessment. Our experimental framework is implemented using PyTorch, with all training performed on an NVIDIA 3090 GPU. The model is configured with 32 feature channels. We employed a cosine-annealed decay strategy for the learning rate, with an initial rate of $5 \times 10^{-4}$ that gradually decays to $5 \times 10^{-8}$ over 500 epochs. We optimize the model using the Adam optimizer~\cite {kingma2014adam}, with gradient clipping set to 4.

\subsection{Comparison with the SOTA Methods}
\begin{table*}[htpb]
\centering
\small  
\renewcommand{\arraystretch}{1.0}
\renewcommand{\tabcolsep}{3pt}  
\caption{Evaluation of our method on real-world full-resolution scenes from three datasets.\label{table_full}}
\resizebox{0.8\linewidth}{!}  
{
\begin{tabular}{c|ccc|ccc|ccc}
\hline
& \multicolumn{3}{c|}{Full-Jilin-Ave}                                        
& \multicolumn{3}{c|}{Full-Landsat-Ave}                                      
& \multicolumn{3}{c}{Full-Skysat-Ave}    
\\ \cline{2-10} 
\multirow{-2}{*}{Method} 
& $D_{\lambda}$ $\downarrow$        & $D_{S}$ $\downarrow$   & QNR $\uparrow$      
& $D_{\lambda}$ $\downarrow$        & $D_{S}$ $\downarrow$   & QNR $\uparrow$    
& $D_{\lambda}$ $\downarrow$        & $D_{S}$ $\downarrow$   & QNR $\uparrow$    \\ \hline
GS & 0.0812 & 0.2021 & 0.7338 & 0.0568 & 0.1110 & 0.8393 & 0.0623 & 0.1112 & 0.8347 \\
IHS & 0.0971 & 0.2124 & 0.7116 & 0.0961 & 0.1233 & 0.7936 & 0.0974 & 0.1318 & 0.7852 \\
GFPCA & 0.0424 & 0.0985 & 0.8633 & 0.0649 & 0.0539 & 0.8850 & 0.0433 & 0.0624 & 0.8976 \\ \hline
MSDCNN & 0.0477 & 0.0420 & 0.9126 & 0.0488 & 0.0326 & 0.9214 & 0.0419 & \textbf{0.0222} & \underline{0.9403} \\
SFINet & 0.0538 & 0.0439 & 0.9050 & 0.0425 & 0.0342 & 0.9255 & 0.1346 & 0.0334 & 0.8379 \\
MSDDN & \underline{0.0415} & 0.0521 & 0.9097 & 0.0336 & 0.0236 & 0.9442 & 0.0412 & 0.0363 & 0.9264 \\
PanFlowNet & 0.0459 & \textbf{0.0382} & \underline{0.9155} & 0.0482 & 0.0384 & 0.9159 & 0.0777 & 0.0332 & 0.8927 \\
HFIN & 0.0424 & 0.0471 & 0.9129 & \underline{0.0240} & \textbf{0.0197} & \underline{0.9562} & 0.0591 & 0.0590 & 0.8865 \\
Pan-mamba & 0.0465 & 0.0426 & 0.9132 & 0.0309 & 0.0736 & 0.8979 & 0.0533 & 0.0311 & 0.9182 \\
ARConv & 0.0417 & 0.0629 & 0.8980 & 0.0257 & \underline{0.0201} & 0.9550 & \textbf{0.0387} & 0.0409 & 0.9225 \\
\rowcolor{gray!20}
Ours & \textbf{0.0409} & \underline{0.0418} & \textbf{0.9156} & \textbf{0.0234} & 0.0228 & \textbf{0.9569} & \underline{0.0405} & \underline{0.0269} & \textbf{0.9408} \\
\hline
\end{tabular}
}
\end{table*}
\begin{figure*}[htpb]
    \centering
    \includegraphics[width=0.9\textwidth]{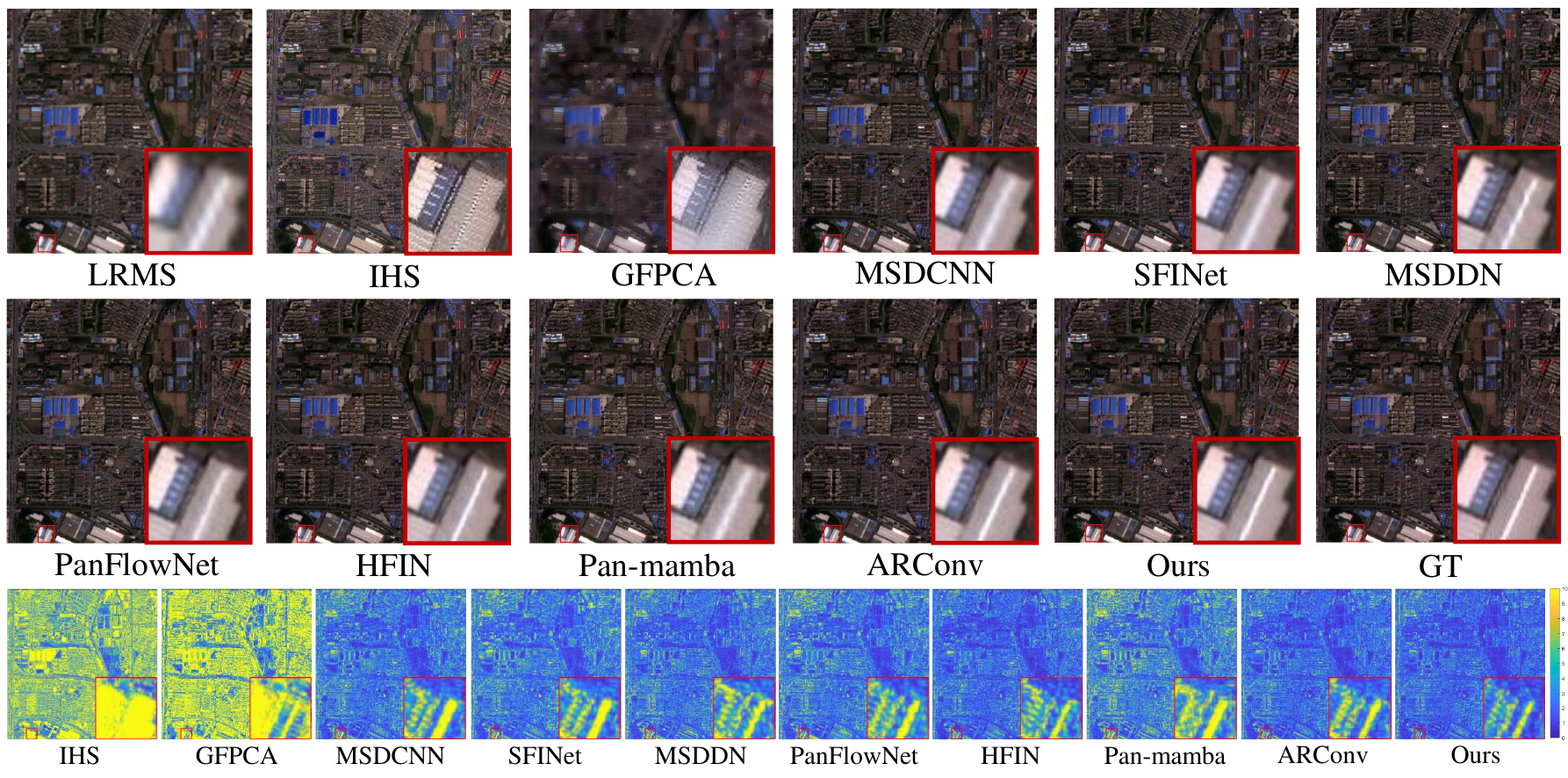}
    \caption{\label{visual_jilin}Comparative visual experiments of several methods on Jilin datasets.}
\end{figure*}
\begin{table*}[!htb]
    \centering
    \small
    \renewcommand{\arraystretch}{1.2}
\renewcommand{\tabcolsep}{3pt}
\caption{Evaluation of parameter count and GFLOPs.}
    \label{table_flops}
\begin{tabular}{c|cccccccc}
\hline
Methods   & MSDCNN & SFINet & MSDDN  & PanFlowNet & HFIN & Pan-mamba  & ARConv  & \cellcolor{gray!20}Ours   \\ \hline
Params(M) & 0.2390 & 0.0484 & 0.3649  & 0.0873  & 1.9836 & 0.1828 & 4.4147    & \cellcolor{gray!20}0.5151 \\ 
GFLOPs(G)  & 9.5488 & 1.8785 & 7.6938 & 13.9469  & 46.2067 & 7.3459 & 38.3152    & \cellcolor{gray!20}20.5679 \\ \hline
\end{tabular}
\end{table*}
\begin{table*}[!htb]
    \centering
    \small
    \renewcommand{\arraystretch}{1.1}
\renewcommand{\tabcolsep}{3pt}
\caption{Ablation study of our method on the Landsat test set. The best are highlighted in \textbf {bold}, and the second are \underline {underlined}.}
    \label{table_abl}
\begin{tabular}{c|cc|cc|cc|cc}
\hline
& \multicolumn{2}{c|}{Landsat-200}   
& \multicolumn{2}{c|}{Landsat-400}
& \multicolumn{2}{c|}{Landsat-800}   
& \multicolumn{2}{c}{Landsat-1600}
\\ \cline{2-9}
\multirow{-2}{*}{Ablation} 
& PSNR$\uparrow$ & SSIM$\uparrow$ 
& PSNR$\uparrow$ & SSIM$\uparrow$ 
& PSNR$\uparrow$ & SSIM$\uparrow$  
& PSNR$\uparrow$ & SSIM$\uparrow$     
\\ \hline
w/o RoPE  & 40.4625 & 0.9682 & 40.9542 & 0.9704  & \underline {40.7619} & 0.9699 & \underline {40.6924}  & 0.9701   \\
SeqT -> SpaT  & \textbf {40.9061} & \textbf {0.9700} & \underline {41.2964} & \underline {0.9724}  & 40.7206 & \underline {0.9714} & 40.5065  & \underline {0.9716}   \\
w/o SAP  & 40.5279 & 0.9660 & 40.9325 & 0.9692  & 40.6186 & 0.9688 & 40.3895  & 0.9689   \\
    \rowcolor{gray!20}
Baseline & \underline {40.6063} & \underline {0.9677} 
& \textbf {41.3715}  & \textbf {0.9725}  & \textbf {41.1326} & \textbf {0.9719} & \textbf {41.0335}  & \textbf {0.9722}   \\ \hline
\end{tabular}
\end{table*}

\textbf{Quantitative Comparison} As shown in the comparative analysis in Table~\ref{table_main}, we evaluated our proposed method against several state-of-the-art pansharpening approaches. For brevity, the reported results represent the average performance across multiple resolution settings within each dataset; the complete results for individual resolutions are provided in the appendix~\ref{append-full-table}. Our method consistently and significantly outperforms all competing methods across all evaluation metrics, demonstrating its strong fusion quality and robustness. To further assess the adaptability of our approach to varying input resolutions, we visualize the PSNR scores of different methods on the Landsat and Skysat datasets across a range of resolution scales, as illustrated in Fig.~\ref{fig_scale}(a) and (b). While our method maintains stable performance as the input resolution increases, the other methods exhibit noticeable performance degradation to varying degrees. These results highlight the superior scalability and generalization ability of ScaleFormer in handling diverse and high-resolution remote sensing data.

To validate the proposed ScaleFormer's real-world performance, we conducted experiments on three full-resolution datasets from the PanScale benchmark, each containing four test sets at varying spatial scales. Due to the lack of ground truth in these real-world scenarios, we employed three no-reference metrics from PanScale-Bench to assess model performance. For conciseness, the main text reports the average results across different resolutions within each dataset, while detailed per-scale results are provided in the appendix~\ref{append-full-table}. As shown in Table~\ref{table_full}, ScaleFormer consistently achieves competitive performance across all evaluation metrics, demonstrating its strong generalization ability and robustness in complex, real-world multi-resolution fusion tasks.

To evaluate the efficiency of the proposed approach, we compare the model's parameter count and computational complexity with those of several baseline methods. All experiments in Table~\ref{table_flops} were conducted with input resolutions set to $200 \times 200$.  ScaleFormer demonstrates a comparable parameter count to Pan-mamba~\cite{he2025panmamba}, which is less than one-fourth of the state-of-the-art methods, HFIN~\cite{tan2024hfin_cvpr24} and ARConv~\cite{wang2025arconv_cvpr25}. This significant reduction in parameter count highlights the efficiency of ScaleFormer relative to these methods. Furthermore, the proposed method exhibits a clear advantage in computational complexity over these two advanced methods, underscoring its competitiveness in resource usage and overall efficiency. As illustrated in Fig.~\ref{fig_scale}(c) and (d), this advantage becomes even more prominent as the input resolution increases, with ScaleFormer maintaining significantly lower GFLOPs and memory consumption compared to SOTA methods. This showcases that ScaleFormer offers a compelling balance between performance and computational cost, making it an efficient solution for large-scale and multi-resolution data.

\noindent\textbf{Qualitative Comparison} The qualitative comparison results are presented in Fig.~\ref{visual_jilin}, with more visualizations in the appendix~\ref{More Visual Comparison}. We visualize representative samples at the largest resolution scale from each dataset, and further magnify selected regions to facilitate detailed comparison. The final row of each qualitative figure shows the mean squared error (MSE) maps between the pansharpened and reference images, where brighter areas indicate larger discrepancies. As shown in these figures, the proposed method demonstrates clear advantages in preserving fine spatial details and spectral fidelity, yielding sharper textures and more visually coherent results.

\subsection{Ablation Studies}
To validate the effectiveness of the proposed method, we conduct a series of ablation studies on the Landsat dataset, focusing on key components of the ScaleFormer architecture. The corresponding PSNR and SSIM results are summarized in Table~\ref{table_abl}. Specifically, we examine the impact of the following modifications: (1) removing the RoPE from the Sequence Transformer (including the single and the cross); (2) replacing the Sequence Transformer with their corresponding Spatial Transformer counterparts, thereby discarding the scale dimension modeling; and (3) removing the SAP by fixing the training window size to a single static setting, aligned with the configuration used during inference.

As shown in the results, all ablation variants exhibit a clear and consistent decline in performance as the test resolution increases, compared to the full model. This degradation highlights the limitations of these configurations in generalizing across scales. In particular, the absence of RoPE hinders the model's capacity to encode relative positional dependencies effectively across varying resolutions. Replacing sequence attention with purely spatial attention deprives the model of its ability to capture inter-patch relationships across sequence-like structures, which are critical for generalizing to unseen scales. Additionally, removing the SAP module prevents the model from learning a diverse and adaptive representation during training, making it less robust to scale shifts encountered at inference time. These findings highlight that the proposed component can enhance scale-awareness and extrapolate beyond the training distribution.

\section{Conclusion}
\label{sec:conclusion}
We propose ScaleFormer, a framework that decouples spatial information from scale-specific features and models them independently. By incorporating a bucketed training strategy and Rotary Position Embeddings (RoPE), ScaleFormer markedly improves scale extrapolation. To enable fair and comprehensive evaluation, we introduce PanScale, addressing the lack of multi-scale inference in existing datasets; to our knowledge, it is the first pansharpening dataset to jointly cover multi-scale and multi-resolution characteristics, accompanied by a tailored evaluation suite, PanScale-Bench. Extensive experiments across PanScale demonstrate that ScaleFormer surpasses state-of-the-art methods in both fusion quality and computational efficiency.
\section{Acknowledgments}
\label{sec:Acknowledgments}
This work was supported by Anhui Provincial Natural Science Foundation under Grant 2408085MD090. We thank Lingting Zhu for the discussions on the method.
{
    \small
    \bibliographystyle{ieeenat_fullname}
    \bibliography{main}
}
\clearpage
\setcounter{page}{1}
\maketitlesupplementary

\section{Related Work}
\label{Appendix Related Work}
\subsection{Pansharpening} 
Pan-sharpening methods can generally be categorized into two major paradigms: traditional techniques and deep learning-based approaches. Traditional methods are further divided into three main categories: component substitution (CS) \cite{GS, SFIM}, multi-resolution analysis (MRA) \cite{mallat1989theory, zhou1998wavelet}, and variational optimization (VO) \cite{ fasbender2008bayesian, li2010new}. CS \cite{gillespie1987color, haydn1982application} techniques enhance spatial resolution by substituting the spatial textures of the LRMS image with high-frequency details extracted from the PAN image. MRA \cite{schowengerdt1980reconstruction, vivone2014critical} approaches perform image information through multi-scale fusion, and VO \cite{ghahremani2015compressed, palsson2013new} methods generate the HRMS by optimizing the loss function. However, traditional methods typically rely on hand-crafted features and lack robust priors, often resulting in suboptimal results. Driven by the success of deep neural networks, deep learning-based methods have subsequently emerged as the predominant approach in the pan-sharpening community \cite{zhou2022mutual,lin2023domain,zhang2024frequency}. The pioneering work of PNN \cite{rs8070594} initiated the application of convolutional neural networks to pan-sharpening. Following this, research efforts have increasingly focused on devising more sophisticated network architectures (e.g., MSDCNN \cite{msdcnn}, GPPNN \cite{gppnn}), incorporating auxiliary information (e.g., frequency analysis in SFINet\cite{zhou2022sfiin} and MSDDN\cite{he2023msddn}), and exploring innovative learning paradigms~\cite{yang2023panflownet,wang2025towards,fang2025pandit}. Concurrently, pan-sharpening has witnessed a significant expansion in its richness and diversity. HFIN \cite{tan2024hfin_cvpr24} introduces a hierarchical frequency integration network. Mamba-based methods~\cite{peng2024fusionmamba,cao2024novel,wang2025self} employ the state-space model to achieve effective pan-sharpening. Furthermore, ARConv \cite{wang2025arconv_cvpr25} presents an adaptive rectangular convolution, enabling the network to learn convolution kernel dimensions tailored to objects of varying scales within remote sensing imagery.

\subsection{Scale Generalization in Perceptual Tasks}
Scale generalization presents a critical challenge for visual models in perceptual tasks, particularly when dealing with inputs of varying sizes and resolutions. The Flexivit~\cite{beyer2023flexivit} was one of the first attempts to address this issue, enabling the model to adapt to inputs with different patch sizes during training, ensuring stable performance across changing scales. Similarly, Resformer~\cite{tian2023resformer} enhances Vision Transformers' generalization ability in tasks with significant scale variations by employing a multi-resolution training strategy with a scale consistency loss function. Building on this, ViTAR~\cite{fan2024vitar} further improves scale generalization by incorporating dynamic resolution adjustments. Additionally, it introduces fuzzy positional encoding to prevent overfitting to a single resolution. The MSPE~\cite{liu2024mspe} method optimizes the patch embedding process with multiple variable-sized patch kernels, pushing the performance of Vision Transformers to new heights, especially in tasks involving multi-scale images, where it demonstrates a substantial improvement in model performance.

\section{PanScale Dataset and Benchmark}
\subsection{Dataset Preprocessing}
\label{Detailed Information for Preprocessing}
\begin{figure*}[h]
    \centering
    \includegraphics[width=\textwidth]{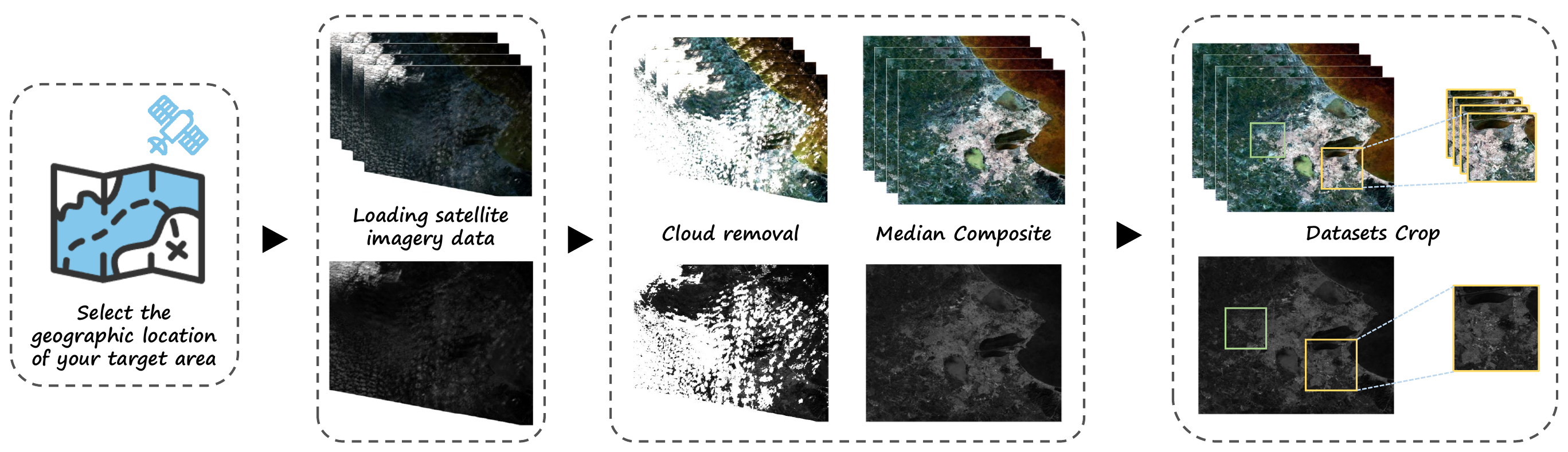}
    \caption{\label{dataset_pipe}The construction pipeline of the PanScale dataset, which consists of three main stages: accessing satellite imagery via Google Earth Engine, data preprocessing, and data cropping.}
\end{figure*}
As is shown in Fig.~\ref{dataset_pipe}, we first selected 16 geographically diverse regions across the globe, covering a broad spectrum of terrains such as urban, mountainous, oceanic, and desert. This diversity ensures that the resulting dataset effectively captures the challenges of pansharpening under varying environmental conditions, thereby facilitating robust evaluation of model generalization across different geographic contexts. Based on this selection, we sourced imagery from three widely used satellites: Jilin-1, Landsat-9, and Skysat. Taking Landsat imagery as an example, each raw satellite image can be represented as $\mathbf{I}\in {{\mathbb{R}}^{H\times W\times B}}$, where $H$,$W$, and $B$ denote the height, width, and number of spectral bands. To remove cloud-contaminated pixels, we utilize the quality assessment band $Q$, which provides metadata indicating the presence of clouds and cirrus. Using this metadata, a binary cloud mask $Mask$ is generated, and the resulting cloud-free image is defined as:
\begin{equation}\label{}
\mathbf{I}'=\mathbf{I}\odot Mask
\end{equation}
Building upon the cloud-free images, we collect all available Landsat satellite observations covering a selected geographic region within a defined time interval $\left[ t_0, t_1 \right]$, resulting in an image set: 
\begin{equation}\label{}
\mathcal{I}=\left\{ {{{\mathbf{I}'}}_{1}},{{{\mathbf{I}'}}_{2}},...,{{{\mathbf{I}'}}_{N}} \right\}
\end{equation}
For each spectral band $b \in [1, B]$, we perform a pixel-wise median compositing across all cloud-free images in the set $\mathcal{I}$ to generate a temporally stable and noise-reduced composite band:
\begin{equation}\label{}
\mathbf{I}_{b}^{mid}(i,j)=median(\{{\mathbf{I}_{1,b}}(i,j),{\mathbf{I}_{2,b}}(i,j),...,{\mathbf{I}_{N,b}}(i,j)\})
\end{equation}
where $(i, j)$ denotes the spatial pixel location with $i \in [1, H]$ and $j \in [1, W]$. By stacking the median composites of all spectral bands, we obtain the final multi-band synthesized image, denoted as $\mathbf{I}^{\text{med}}$. Leveraging the band extraction and export capabilities of Google Earth Engine, we further derive the corresponding panchromatic (PAN) images and low-resolution multi-spectral (LRMS) images for subsequent use in pansharpening.
\subsection{Dataset Construction}
\label{Detailed Information for Dataset Construction}
Table~\ref{table_panscale} presents the detailed configuration of the dataset construction. The training sets for the three constituent sources—Jilin-1, Landsat-9, and Skysat—are designed with distinct upsampling ratios of 4$\times$, 2$\times$, and 2.5$\times$, respectively. This variation in scaling allows the models to be exposed to diverse resolution levels during training.

For a comprehensive evaluation, we establish a multi-resolution testing framework to rigorously assess the cross-scale generalization ability of pansharpening models. Specifically, the reduced-resolution (RR) testing sets consist of images at resolutions of $1600 \times 1600$, $800 \times 800$, $400 \times 400$, and $200 \times 200$ pixels. Notably, for the Jilin dataset, the maximum available resolution is 800×800, and therefore, the $1600 \times 1600$ variant is not included. In addition to the RR evaluations, we provide full-resolution (FR) test sets at varying scales to further evaluate model robustness across real-world spatial variations.
\begin{table*}[!h]
    \centering
    \small
    \renewcommand{\arraystretch}{1.2}
    \renewcommand{\tabcolsep}{3pt}
    \caption{Detailed Information of the PanScale Datasets, where "samples" denotes the number of samples at each scale, "scale" represents the height or width of the square PAN image, and "res" indicates the image resolution used for dataset creation.}
    \label{table_panscale}
    \begin{tabular}{c|ccc|ccc|ccc}
    \hline
    \rowcolor{gray!20}
    & \multicolumn{3}{c|}{\textbf{Jilin (4$\times$)}}   
    & \multicolumn{3}{c|}{\textbf{Landsat (2$\times$)}} 
    & \multicolumn{3}{c}{\textbf{Skysat (2.5$\times$)}} \\ \cline{2-10}
    \rowcolor{gray!10} 
    & \textbf{Samples} & \textbf{Scale} & \textbf{Res(m)} 
    & \textbf{Samples} & \textbf{Scale} & \textbf{Res(m)} 
    & \textbf{Samples} & \textbf{Scale} & \textbf{Res(m)} \\ \hline
    Train     
    & 1055  & 200           & 2  
    & 2484  & 256           & 30 
    & 2370  & 200           & 2  \\ \hline
    RR Test    
    & 34    & \makecell{800,400 \\ 200}    & 2  
    & 96    & \makecell{1600,800 \\ 400,200} & 30 
    & 88    & \makecell{1600,800 \\ 400,200} & 2  \\ \hline
    FR Test    
    & 121   & \makecell{2000,1600 \\ 1200,800} & 2  
    & 48    & \makecell{2000,1600 \\ 1200,800} & 30 
    & 34    & \makecell{2000,1600 \\ 1200,800} & 2  \\ \hline
    \end{tabular}
\end{table*}

\subsection{Benchmark Design}
\label{Appendix Benchmark Design}
To comprehensively evaluate diverse methods on the PanScale benchmark, we employ a suite of image quality assessment (IQA) metrics. Specifically, we select six prevalent reference-based metrics: Peak Signal-to-Noise Ratio (PSNR) \cite{huynh2008scope}, Structural Similarity Index (SSIM) \cite{wang2004image}, Spectral Angle Mapper (SAM) \cite{yuhas1992discrimination}, Relative Dimensionless Global Error in Synthesis (ERGAS) \cite{wald2002data}, Spatial Correlation Coefficient (SCC) \cite{kelejian1995spatial} and Quality index (Q) \cite{vivone2021benchmarking}. PSNR and SSIM primarily quantify spatial fidelity, with higher values indicating superior preservation of spatial details.  Conversely, SAM and ERGAS focus on assessing spectral accuracy, where lower values denote diminished spectral distortion and enhanced consistency with the reference data. SCC assesses the spatial correlation between images, while Q provides a comprehensive measure of universal image quality for pansharpened MS images. Furthermore, to provide additional evaluation of model performance on PanScale, we also leverage three reference-free IQA metrics that do not rely on ground truth data. These include the spectral distortion index $D_{\lambda}$ \cite{khan2009pansharpening}, the spatial distortion index $D_{S}$, and the Quality with No Reference (QNR) \cite{alparone2008multispectral}. Specifically, $D_{\lambda}$ and $D_{S}$ quantify the extent of spectral and spatial distortion, respectively. QNR provides a holistic assessment of image quality without reference.

\section{Additional experimental results}
\subsection{More Visual Comparison}
\label{More Visual Comparison}
\begin{figure*}[h]
    \centering
    \includegraphics[width=\textwidth]{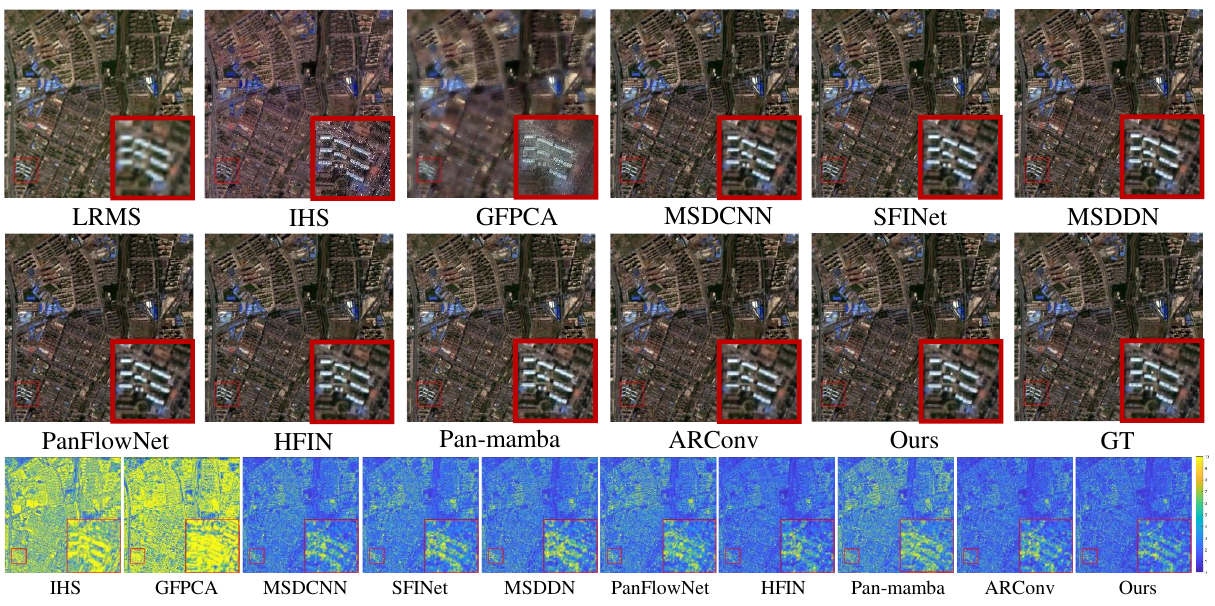}
    \caption{\label{visual_jilin2}Comparative visual experiments of several methods on Jilin datasets.}
\end{figure*}

\begin{figure*}[h]
    \centering
    \includegraphics[width=\textwidth]{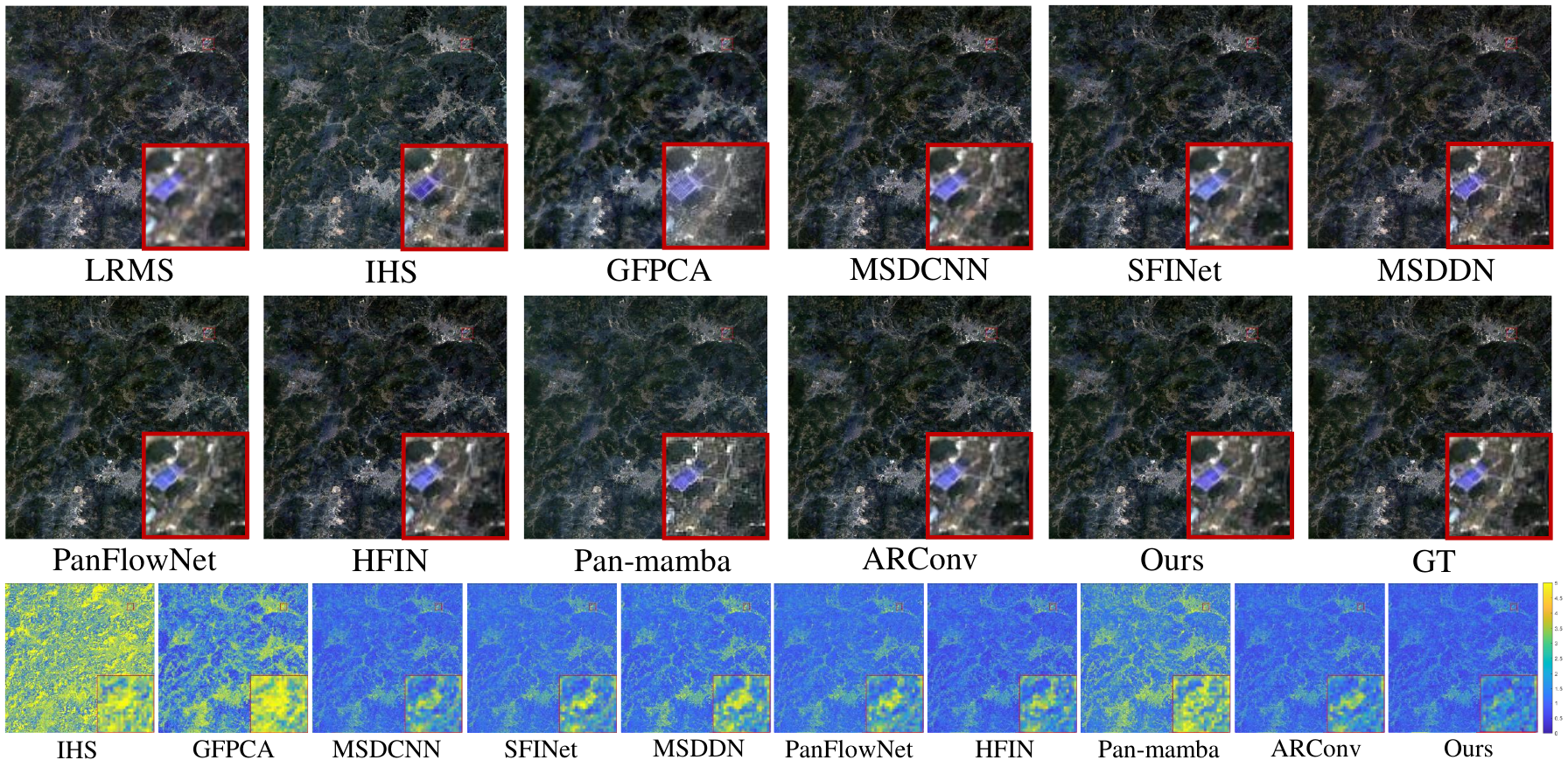}
    \caption{\label{visual_landsat}Comparative visual experiments of several methods on Landsat datasets.}
\end{figure*}

\begin{figure*}[h]
    \centering
    \includegraphics[width=\textwidth]{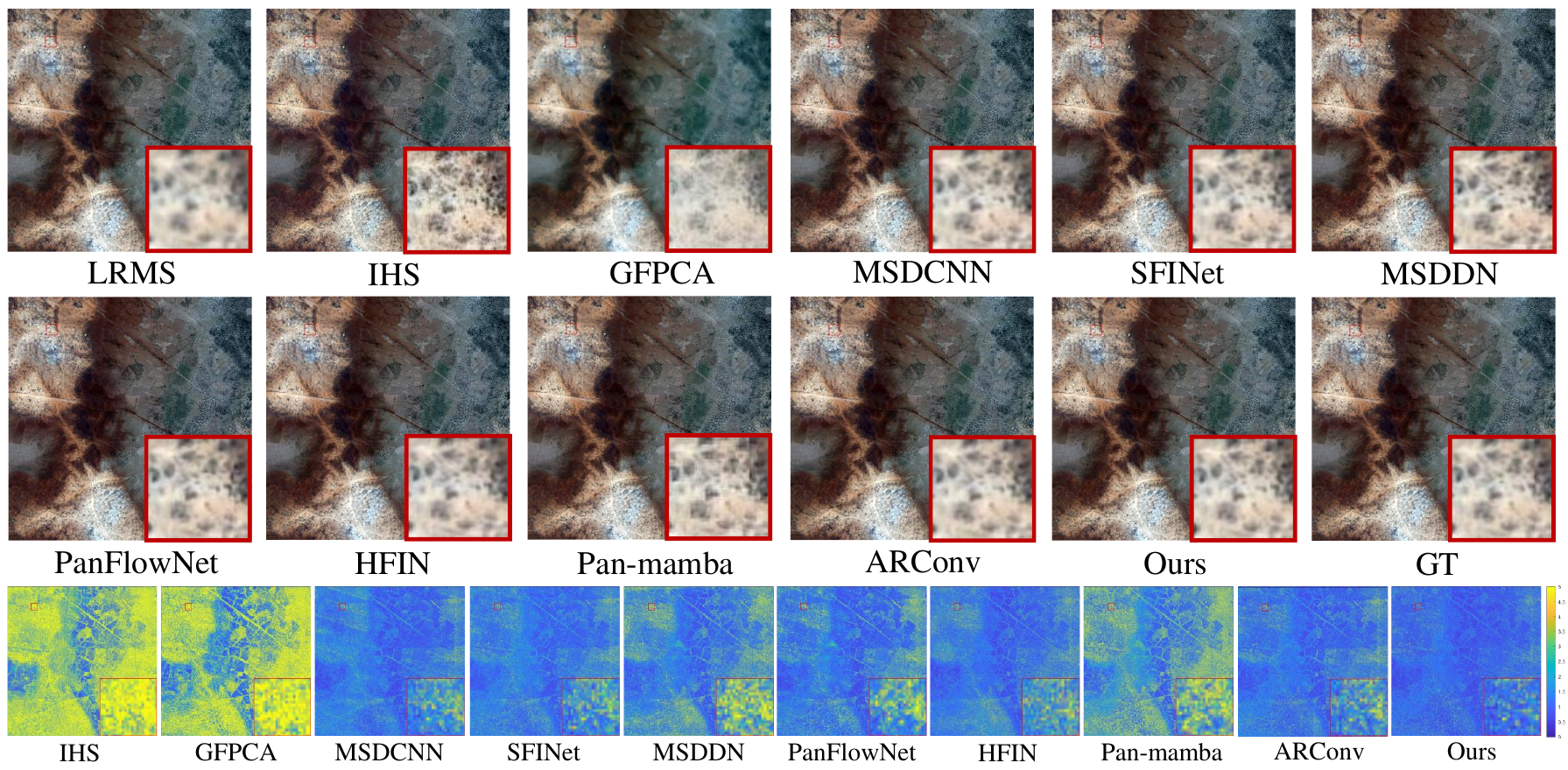}
    \caption{\label{visual_skysat}Comparative visual experiments of several methods on Skysat datasets.}
\end{figure*}

Due to space constraints, we provide only the visualization results for the Jilin dataset in the main text. More visual comparison results for the three datasets are shown in Fig.~\ref{visual_jilin2}, ~\ref{visual_landsat}, and ~\ref {visual_skysat}. The zoomed-in comparison images and corresponding MSE maps demonstrate that the proposed method excels in preserving fine details and spectral fidelity. These results highlight its strong adaptability to large-scale pansharpening tasks, showcasing its ability to handle high-resolution image fusion effectively.
\subsection{Comprehensive Quantitative Comparison}
\label{append-full-table}
Due to space limitations, we report only the average multi-scale performance on the reduced resolution datasets in Table~\ref{table_main} of the main text. For completeness and reproducibility, the full quantitative results across all resolutions and datasets are provided in the following tables: Table~\ref{table_jilin_all_part1} (Jilin dataset at 200 and 400 resolutions), Table~\ref{table_jilin_all_part2} (Jilin dataset at 800 resolution), Table~\ref{table_landsat_all_part1} (Landsat dataset at 200 and 400 resolutions), Table~\ref{table_landsat_all_part2} (Landsat dataset at 800 and 1600 resolutions), Table~\ref{table_skysat_all_part1} (Skysat dataset at 200 and 400 resolutions), and Table~\ref{table_skysat_all_part2} (Skysat dataset at 800 and 1600 resolutions). These results offer a more detailed perspective on the performance of each method across different scales. Notably, our proposed method consistently outperforms existing approaches across nearly all metrics, demonstrating its superior cross-scale generalization capability.

Similarly, for the full-resolution experiments summarized in Table~\ref{table_full} of the main text, we report only the averaged results across multiple scales. To ensure completeness and facilitate further analysis, the detailed performance metrics for each resolution are presented in the following supplementary tables: Table~\ref{table_full_jilin_all} (Full Jilin dataset at 800, 1200, 1600, and 2000 resolutions), Table~\ref{table_full_landsat_all} (Full Landsat dataset at 800, 1200, 1600, and 2000 resolutions), and Table~\ref{table_full_skysat_all} (Full Skysat dataset at 800, 1200, 1600, and 2000 resolutions). These comprehensive results show that our method achieves performance on par with or superior to state-of-the-art techniques even under full-resolution settings, demonstrating strong generalization capability across scales in real-world scenarios.
\begin{table*}[htpb]
\centering
\small  
\renewcommand{\arraystretch}{1.1}
\renewcommand{\tabcolsep}{3pt}  
\caption{Multi-resolution results on the Jilin datasets (Part 1), with numbers indicating the size of the PAN image. The best are highlighted in \textbf {bold}.\label{table_jilin_all_part1}}
\resizebox{\linewidth}{!}
{
\begin{tabular}{c|cccccc|cccccc}
\hline
& \multicolumn{6}{c|}{Jilin-200}  
& \multicolumn{6}{c}{Jilin-400}
\\ \cline{2-13} 
\multirow{-2}{*}{Method} 
& PSNR$\uparrow$      & SSIM$\uparrow$    & SAM$\downarrow$ 
& ERGAS$\downarrow$   & SCC $\uparrow$    & Q $\uparrow$ 
& PSNR$\uparrow$      & SSIM$\uparrow$    & SAM$\downarrow$ 
& ERGAS$\downarrow$   & SCC $\uparrow$    & Q $\uparrow$ \\ \hline
GS & 25.7681 & 0.7128 & 0.0573 & 5.3592 & 0.8592 & 0.6002 & 25.6914 & 0.7099 & 0.0570 & 5.5151 & 0.8683 & 0.5805 \\
IHS & 25.9327 & 0.7081 & 0.0590 & 5.6410 & 0.8488 & 0.5930 & 25.8671 & 0.7063 & 0.0588 & 5.7120 & 0.8618 & 0.5754 \\
GFPCA & 25.6482 & 0.7106 & 0.0761 & 5.1018 & 0.8730 & 0.4579 & 25.8554 & 0.7252 & 0.0733 & 4.9484 & 0.8882 & 0.4469 \\ \hline
MSDCNN & 36.5984 & 0.9555 & 0.0277 & 1.5336 & 0.9885 & 0.9146 & 37.0869 & 0.9588 & 0.0265 & 1.4275 & 0.9907 & 0.9056 \\
SFINet & 36.2407 & 0.9508 & 0.0306 & 1.6494 & 0.9872 & 0.9017 & 36.6916 & 0.9532 & 0.0301 & 1.5592 & 0.9895 & 0.8915 \\
MSDDN & 37.5748 & 0.9653 & 0.0242 & 1.3600 & 0.9910 & 0.9289 & 37.7031 & 0.9667 & 0.0244 & 1.3366 & 0.9922 & 0.9174 \\
PanFlowNet & 36.6522 & 0.9571 & 0.0277 & 1.5303 & 0.9886 & 0.9163 & 36.8303 & 0.9583 & 0.0272 & 1.4832 & 0.9903 & 0.9038 \\
HFIN & 38.0366 & 0.9686 & 0.0228 & 1.3103 & 0.9919 & 0.9343 & 38.2695 & 0.9693 & 0.0224 & 1.2464 & 0.9932 & 0.9223 \\
Pan-mamba & 36.1465 & 0.9517 & 0.0278 & 1.6057 & 0.9873 & 0.9092 & 36.3710 & 0.9534 & 0.0280 & 1.5434 & 0.9891 & 0.8985 \\
ARConv & 38.1115 & 0.9682 & 0.0233 & 1.2904 & 0.9919 & 0.9338 & 38.5962 & 0.9701 & 0.0225 & 1.2113 & 0.9934 & 0.9246 \\
\rowcolor{gray!20}
Ours & \textbf {39.3300} & \textbf {0.9759} & \textbf {0.0202} & \textbf {1.1152} & \textbf {0.9941} & \textbf {0.9449} & \textbf {39.6334} & \textbf {0.9762} & \textbf {0.0197} & \textbf {1.0728} & \textbf {0.9950} & \textbf {0.9336} \\
\hline
\end{tabular}
}
\end{table*}

\begin{table*}[htpb]
\centering
\small  
\renewcommand{\arraystretch}{1.1}
\renewcommand{\tabcolsep}{3pt}  
\caption{Multi-resolution results on the Jilin datasets (Part 2), with numbers indicating the size of the PAN image. The best are highlighted in \textbf {bold}.\label{table_jilin_all_part2}}
\resizebox{0.5\linewidth}{!}
{
\begin{tabular}{c|cccccc}
\hline
& \multicolumn{6}{c}{Jilin-800}  
\\ \cline{2-7} 
\multirow{-2}{*}{Method} 
& PSNR$\uparrow$      & SSIM$\uparrow$    & SAM$\downarrow$ 
& ERGAS$\downarrow$   & SCC $\uparrow$    & Q $\uparrow$ \\ \hline
GS & 24.8563 & 0.6939 & 0.0605 & 5.9530 & 0.8700 & 0.5739 \\
IHS & 24.9852 & 0.6926 & 0.0611 & 6.0682 & 0.8659 & 0.5726 \\
GFPCA & 25.3367 & 0.7198 & 0.0738 & 5.0877 & 0.8974 & 0.4479 \\ \hline
MSDCNN & 36.0279 & 0.9578 & 0.0263 & 1.5284 & 0.9907 & 0.9074 \\
SFINet & 35.9211 & 0.9528 & 0.0295 & 1.6301 & 0.9901 & 0.8942 \\
MSDDN & 35.5442 & 0.9617 & 0.0266 & 1.6012 & 0.9903 & 0.9122 \\
PanFlowNet & 35.5154 & 0.9543 & 0.0278 & 1.6527 & 0.9897 & 0.9001 \\
HFIN & 37.6967 & 0.9714 & 0.0219 & 1.2696 & 0.9940 & 0.9285 \\
Pan-mamba & 34.1367 & 0.9388 & 0.0311 & 1.9399 & 0.9858 & 0.8822 \\
ARConv & 37.9942 & 0.9708 & 0.0221 & 1.2368 & 0.9940 & 0.9288 \\
\rowcolor{gray!20}
Ours & \textbf {38.9161} & \textbf {0.9762} & \textbf {0.0196} & \textbf {1.1093} & \textbf {0.9953} & \textbf {0.9369} \\
\hline
\end{tabular}
}
\end{table*}

\begin{table*}[htpb]
\centering
\small  
\renewcommand{\arraystretch}{1.1}
\renewcommand{\tabcolsep}{3pt}  
\caption{Multi-resolution results on the Landsat datasets (Part 1), with numbers indicating the size of the PAN image. The best are highlighted in \textbf {bold}.\label{table_landsat_all_part1}}
\resizebox{\linewidth}{!}
{
\begin{tabular}{c|cccccc|cccccc}
\hline
& \multicolumn{6}{c|}{Landsat-200}  
& \multicolumn{6}{c}{Landsat-400}
\\ \cline{2-13} 
\multirow{-2}{*}{Method} 
& PSNR$\uparrow$      & SSIM$\uparrow$    & SAM$\downarrow$ 
& ERGAS$\downarrow$   & SCC $\uparrow$    & Q $\uparrow$ 
& PSNR$\uparrow$      & SSIM$\uparrow$    & SAM$\downarrow$ 
& ERGAS$\downarrow$   & SCC $\uparrow$    & Q $\uparrow$ \\ \hline
GS & 36.6649 & 0.9049 & 0.0413 & 4.1725 & 0.8227 & 0.5872 & 36.1124 & 0.8994 & 0.0481 & 4.0663 & 0.8504 & 0.5853 \\
IHS & 37.3551 & 0.9001 & 0.0364 & 4.1847 & 0.7969 & 0.5566 & 36.7972 & 0.8952 & 0.0398 & 4.0051 & 0.8263 & 0.5547 \\
GFPCA & 37.1018 & 0.8968 & 0.0432 & 3.7622 & 0.8255 & 0.4707 & 36.9330 & 0.8975 & 0.0431 & 3.3999 & 0.8967 & 0.4744 \\ \hline
MSDCNN & 38.8386 & 0.9590 & 0.0407 & 2.8789 & 0.8242 & 0.6667 & 38.6542 & 0.9597 & 0.0396 & 2.7907 & 0.8442 & 0.6678 \\
SFINet & 38.7072 & 0.9523 & 0.0486 & 3.3100 & 0.8428 & 0.6666 & 39.1371 & 0.9569 & 0.0484 & 2.7937 & 0.8649 & 0.6683 \\
MSDDN & 37.8541 & 0.9360 & 0.0609 & 3.2539 & 0.8212 & 0.6442 & 38.3405 & 0.9391 & 0.0524 & 2.7208 & 0.8213 & 0.6433 \\
PanFlowNet & 37.9274 & 0.9343 & 0.0721 & 3.8958 & 0.8263 & 0.6571 & 38.3768 & 0.9423 & 0.0627 & 3.3714 & 0.8470 & 0.6592 \\
HFIN & 39.9710 & 0.9635 & 0.0345 & 2.8424 & 0.8768 & 0.7165 & 40.6129 & 0.9698 & 0.0337 & 2.2585 & 0.9031 & 0.7208 \\
Pan-mamba & 39.505 & 0.9553 & 0.0530 & 3.0258 & 0.8701 & 0.7093 & 38.7678 & 0.9471 & 0.0487 & 2.6194 & 0.8512 & 0.6703 \\
ARConv & 39.6267 & 0.9624 & 0.0365 & 2.8184 & 0.8411 & 0.6770 & 39.8234 & 0.9645 & 0.0368 & 2.4587 & 0.8713 & 0.6812 \\
\rowcolor{gray!20}
Ours & \textbf {40.6063} & \textbf {0.9677} & \textbf {0.0341} & \textbf {2.6561} & \textbf {0.8869} & \textbf {0.7317} & \textbf {41.3715} & \textbf {0.9725} & \textbf {0.0333} & \textbf {2.1051} & \textbf {0.9081} & \textbf {0.7367} \\
\hline
\end{tabular}
}
\end{table*}

\begin{table*}[htpb]
\centering
\small  
\renewcommand{\arraystretch}{1.1}
\renewcommand{\tabcolsep}{3pt}  
\caption{Multi-resolution results on the Landsat datasets (Part 2), with numbers indicating the size of the PAN image. The best are highlighted in \textbf {bold}.\label{table_landsat_all_part2}}
\resizebox{\linewidth}{!}
{
\begin{tabular}{c|cccccc|cccccc}
\hline
& \multicolumn{6}{c|}{Landsat-800}  
& \multicolumn{6}{c}{Landsat-1600}
\\ \cline{2-13} 
\multirow{-2}{*}{Method} 
& PSNR$\uparrow$      & SSIM$\uparrow$    & SAM$\downarrow$ 
& ERGAS$\downarrow$   & SCC $\uparrow$    & Q $\uparrow$ 
& PSNR$\uparrow$      & SSIM$\uparrow$    & SAM$\downarrow$ 
& ERGAS$\downarrow$   & SCC $\uparrow$    & Q $\uparrow$ \\ \hline
GS & 34.9034 & 0.8949 & 0.0517 & 4.6372 & 0.8873 & 0.5911 & 32.6318 & 0.8817 & 0.0633 & 5.5056 & 0.9064 & 0.5857 \\
IHS & 35.6661 & 0.8915 & 0.0407 & 4.2507 & 0.8640 & 0.5616 & 33.6923 & 0.8791 & 0.0461 & 4.8532 & 0.8755 & 0.5486 \\
GFPCA & 36.1215 & 0.8963 & 0.0413 & 3.5643 & 0.9074 & 0.4790 & 34.8823 & 0.8946 & 0.0418 & 3.6901 & 0.9320 & 0.4748 \\ \hline
MSDCNN & 38.7256 & 0.9600 & 0.0388 & 2.6008 & 0.8973 & 0.6809 & 39.0508 & 0.9582 & 0.0430 & 2.3577 & 0.9553 & 0.6792 \\
SFINet & 39.1135 & 0.9575 & 0.0461 & 2.5864 & 0.9167 & 0.6804 & 38.3526 & 0.9600 & 0.0389 & 2.5399 & 0.9314 & 0.6803 \\
MSDDN & 37.7455 & 0.9364 & 0.0506 & 2.7781 & 0.8704 & 0.6480 & 36.8196 & 0.9269 & 0.0515 & 2.8563 & 0.9072 & 0.6329 \\
PanFlowNet & 38.2522 & 0.9438 & 0.0573 & 3.1225 & 0.8994 & 0.6735 & 38.3067 & 0.9527 & 0.0475 & 2.6904 & 0.9453 & 0.6777 \\
HFIN & 40.2656 & 0.9664 & 0.0343 & 2.1820 & 0.9429 & 0.7249 & 39.9946 & 0.9665 & 0.0349 & 2.0090 & 0.9701 & 0.7243 \\
Pan-mamba & 35.7809 & 0.9131 & 0.0579 & 3.2307 & 0.8508 & 0.6291 & 32.8549 & 0.8668 & 0.0744 & 4.1172 & 0.8376 & 0.5661 \\
ARConv & 39.7381 & 0.9642 & 0.0342 & 2.3293 & 0.9185 & 0.6943 & 39.4700 & 0.9642 & 0.0339 & 2.1698 & 0.9511 & 0.6939 \\
\rowcolor{gray!20}
Ours & \textbf {41.1326} & \textbf {0.9719} & \textbf {0.0327} & \textbf {1.9870} & \textbf {0.9471} & \textbf {0.7438} & \textbf {41.0335} & \textbf {0.9722} & \textbf {0.0317} & \textbf {1.7934} & \textbf {0.9735} & \textbf {0.7458} \\
\hline
\end{tabular}
}
\end{table*}

\begin{table*}[htpb]
\centering
\small  
\renewcommand{\arraystretch}{1.1}
\renewcommand{\tabcolsep}{3pt}  
\caption{Multi-resolution results on the Skysat datasets (Part 1), with numbers indicating the size of the PAN image. The best are highlighted in \textbf {bold}.\label{table_skysat_all_part1}}
\resizebox{\linewidth}{!}
{
\begin{tabular}{c|cccccc|cccccc}
\hline
& \multicolumn{6}{c|}{Skysat-200}  
& \multicolumn{6}{c}{Skysat-400}
\\ \cline{2-13} 
\multirow{-2}{*}{Method} 
& PSNR$\uparrow$      & SSIM$\uparrow$    & SAM$\downarrow$ 
& ERGAS$\downarrow$   & SCC $\uparrow$    & Q $\uparrow$ 
& PSNR$\uparrow$      & SSIM$\uparrow$    & SAM$\downarrow$ 
& ERGAS$\downarrow$   & SCC $\uparrow$    & Q $\uparrow$ \\ \hline
GS & 39.4693 & 0.9354 & 0.0326 & 2.2604 & 0.9121 & 0.6343 & 39.2596 & 0.9346 & 0.0310 & 2.3096 & 0.9357 & 0.6478 \\
IHS & 40.1768 & 0.9354 & \textbf {0.0264} & 2.0909 & 0.8934 & 0.6259 & 39.7388 & 0.9326 & \textbf {0.0260} & 2.1468 & 0.9206 & 0.6362 \\
GFPCA & 38.0562 & 0.9289 & 0.0454 & 2.4658 & 0.8896 & 0.4720 & 37.8788 & 0.9275 & 0.0441 & 2.4482 & 0.9213 & 0.4867 \\ \hline
MSDCNN & 43.0041 & 0.9758 & 0.0406 & 2.8975 & 0.8886 & 0.7373 & 43.1001 & 0.9788 & 0.0399 & 2.2283 & 0.9259 & 0.7450 \\
SFINet & 41.6073 & 0.9417 & 0.1010 & 3.4414 & 0.8496 & 0.6763 & 41.3735 & 0.9418 & 0.0939 & 3.1682 & 0.8911 & 0.6829 \\
MSDDN & 42.3969 & 0.9462 & 0.0825 & 2.9244 & 0.8947 & 0.7115 & 41.7617 & 0.9489 & 0.0697 & 2.7420 & 0.9243 & 0.7096 \\
PanFlowNet & 41.0920 & 0.9484 & 0.0757 & 3.2361 & 0.8467 & 0.6674 & 40.9916 & 0.9508 & 0.0667 & 2.9674 & 0.9009 & 0.6703 \\
HFIN & 43.7949 & 0.9533 & 0.0749 & 2.6743 & 0.9073 & 0.7637 & 43.9910 & 0.9669 & 0.0561 & 2.2725 & 0.9443 & 0.7758 \\
Pan-mamba & 43.2983 & 0.9516 & 0.0908 & 2.8129 & 0.8955 & 0.7546 & 42.7551 & 0.9590 & 0.0679 & 2.5895 & 0.9298 & 0.7461 \\
ARConv & 43.5032 & 0.9785 & 0.0423 & 2.1815 & 0.8934 & 0.7475 & 43.4617 & 0.9795 & 0.0406 & 2.0500 & 0.9344 & 0.7596 \\
\rowcolor{gray!20}
Ours & \textbf {44.9275} & \textbf {0.9826} & 0.0362 & \textbf {1.7945} & \textbf {0.9339} & \textbf {0.7837} & \textbf {44.6718} & \textbf {0.9824} & 0.0364 & \textbf {1.7266} & \textbf {0.9519} & \textbf {0.7874} \\
\hline
\end{tabular}
}
\end{table*}

\begin{table*}[htpb]
\centering
\small  
\renewcommand{\arraystretch}{1.1}
\renewcommand{\tabcolsep}{3pt}  
\caption{Multi-resolution results on the Skysat datasets (Part 2), with numbers indicating the size of the PAN image. The best are highlighted in \textbf {bold}.\label{table_skysat_all_part2}}
\resizebox{\linewidth}{!}
{
\begin{tabular}{c|cccccc|cccccc}
\hline
& \multicolumn{6}{c|}{Skysat-800}  
& \multicolumn{6}{c}{Skysat-1600}
\\ \cline{2-13} 
\multirow{-2}{*}{Method} 
& PSNR$\uparrow$      & SSIM$\uparrow$    & SAM$\downarrow$ 
& ERGAS$\downarrow$   & SCC $\uparrow$    & Q $\uparrow$ 
& PSNR$\uparrow$      & SSIM$\uparrow$    & SAM$\downarrow$ 
& ERGAS$\downarrow$   & SCC $\uparrow$    & Q $\uparrow$ \\ \hline
GS & 38.9661 & 0.9302 & 0.0322 & 2.4344 & 0.9474 & 0.6367 & 38.5282 & 0.9264 & 0.0328 & 2.4648 & 0.9555 & 0.6294 \\
IHS & 39.3872 & 0.9280 & \textbf {0.0275} & 2.2406 & 0.9361 & 0.6247 & 39.0572 & 0.9255 & \textbf {0.0269} & 2.2241 & 0.9468 & 0.6217 \\
GFPCA & 37.4581 & 0.9282 & 0.0457 & 2.5427 & 0.9345 & 0.4751 & 37.2721 & 0.9272 & 0.0446 & 2.5326 & 0.9442 & 0.4755 \\ \hline
MSDCNN & 43.0467 & 0.9792 & 0.0391 & 2.1924 & 0.9347 & 0.7424 & 40.9681 & 0.9444 & 0.0851 & 3.0750 & 0.9211 & 0.6798 \\
SFINet & 41.3829 & 0.9452 & 0.0813 & 2.9911 & 0.9136 & 0.6787 & 42.6192 & 0.9786 & 0.0391 & 2.3726 & 0.9423 & 0.7419 \\
MSDDN & 40.7615 & 0.9475 & 0.0652 & 2.7640 & 0.9327 & 0.6900 & 39.2365 & 0.9397 & 0.0691 & 2.9878 & 0.9323 & 0.6684 \\
PanFlowNet & 41.1307 & 0.9558 & 0.0609 & 2.7860 & 0.9215 & 0.6733 & 40.8734 & 0.9570 & 0.0598 & 2.7015 & 0.9393 & 0.6801 \\
HFIN & 44.2075 & 0.9714 & 0.0508 & 2.0866 & 0.9581 & 0.7725 & 43.8435 & 0.9716 & 0.0516 & 2.0618 & 0.9680 & 0.7709 \\
Pan-mamba & 40.9513 & 0.9527 & 0.0626 & 2.6565 & 0.9320 & 0.7017 & 38.5358 & 0.9340 & 0.0689 & 3.0077 & 0.9245 & 0.6491 \\
ARConv & 43.4413 & 0.9803 & 0.0398 & 2.0106 & 0.9470 & 0.7594 & 43.2087 & 0.9806 & 0.0391 & 2.0094 & 0.9529 & 0.7591 \\
\rowcolor{gray!20}
Ours & \textbf {44.7328} & \textbf {0.9830} & 0.0356 & \textbf {1.6606} & \textbf {0.9613} & \textbf {0.7819} & \textbf {44.6416} & \textbf {0.9829} & 0.0347 & \textbf {1.7355} & \textbf {0.9686} & \textbf {0.7811}\\
\hline
\end{tabular}
}
\end{table*}

\begin{table*}[htpb]
\centering
\small  
\renewcommand{\arraystretch}{1.1}
\renewcommand{\tabcolsep}{3pt}  
\caption{Multi-resolution results on the Full-Jilin datasets, with numbers indicating the size of the PAN image. The best are highlighted in \textbf {bold}, and the second are \underline {underlined}.\label{table_full_jilin_all}}
\resizebox{\linewidth}{!}  
{
\begin{tabular}{c|ccc|ccc|ccc|ccc}
\hline
& \multicolumn{3}{c|}{Full-Jilin-800}                        & \multicolumn{3}{c|}{Full-Jilin-1200}                       
& \multicolumn{3}{c|}{Full-Jilin-1600}    
& \multicolumn{3}{c}{Full-Jilin-2000} 
\\ \cline{2-13} 
\multirow{-2}{*}{Method} 
& $D_{\lambda}$ $\downarrow$        & $D_{S}$ $\downarrow$   & QNR $\uparrow$      
& $D_{\lambda}$ $\downarrow$        & $D_{S}$ $\downarrow$   & QNR $\uparrow$    
& $D_{\lambda}$ $\downarrow$        & $D_{S}$ $\downarrow$   & QNR $\uparrow$    
& $D_{\lambda}$ $\downarrow$        & $D_{S}$ $\downarrow$   & QNR $\uparrow$ \\ \hline
GS & 0.0795 & 0.1936 & 0.7433 & 0.0791 & 0.2012 & 0.7362 & 0.0814 & 0.2055 & 0.7304 & 0.0846 & 0.2081 & 0.7253 \\
IHS & 0.0955 & 0.2060 & 0.7191 & 0.0961 & 0.2121 & 0.7126 & 0.0972 & 0.2150 & 0.7092 & 0.0997 & 0.2165 & 0.7056 \\
GFPCA & $\underline{0.0464}$ & 0.1001 & 0.8584 & 0.0420 & 0.0980 & 0.8641 & 0.0405 & 0.0980 & 0.8654 & 0.0407 & 0.0979 & 0.8654 \\ \hline
MSDCNN & 0.0537 & 0.0499 & 0.8995 & 0.0460 & \underline{0.0416} & 0.9144 & 0.0455 & \underline{0.0394} & 0.9170 & 0.0454 & 0.0371 & 0.9194 \\
SFINet & 0.0596 & 0.0515 & 0.8925 & 0.0519 & 0.0430 & 0.9075 & 0.0517 & 0.0415 & 0.9092 & 0.0519 & 0.0395 & 0.9109 \\
MSDDN & 0.0481 & 0.0521 & \underline{0.9046} & \underline{0.0398} & 0.0509 & 0.9118 & \textbf{0.0387} & 0.0521 & 0.9116 & 0.0394 & 0.0533 & 0.9106 \\
PanFlowNet & 0.0502 & \textbf{0.0438} & 0.9044 & 0.0455 & \textbf{0.0384} & \underline{0.9179} & 0.0441 & \textbf{0.0364} & 0.9201 & 0.0439 & \textbf{0.0340} & \textbf{0.9196} \\
HFIN & 0.0479 & 0.0522 & 0.9030 & 0.0414 & 0.0470 & 0.9138 & 0.0403 & 0.0453 & 0.9164 & 0.0398 & 0.0437 & 0.9182 \\
Pan-mamba & 0.0516 & 0.0501 & 0.9013 & 0.0446 & 0.0427 & 0.9148 & 0.0447 & 0.0402 & \underline{0.9171} & 0.0450 & 0.0373 & \underline{0.9195} \\
ARConv & 0.0468 & 0.0656 & 0.8907 & 0.0407 & 0.0629 & 0.8989 & 0.0398 & 0.0619 & 0.9007 & \underline{0.0396} & 0.0611 & 0.9016 \\
\rowcolor{gray!20}
Ours & \textbf{0.0454}& \underline{0.0476} & \textbf{0.9077} & \textbf{0.0394} & 0.0428 & \textbf{0.9181} & \underline{0.0394} & 0.0405 & \textbf{0.9177} & \textbf{0.0393} & \underline{0.0363} & 0.9189 \\
\hline
\end{tabular}
}
\end{table*}

\begin{table*}[htpb]
\centering
\small  
\renewcommand{\arraystretch}{1.1}
\renewcommand{\tabcolsep}{3pt}  
\caption{Multi-resolution results on the Full-Landsat datasets, with numbers indicating the size of the PAN image. The best are highlighted in \textbf {bold}, and the second are \underline {underlined}.\label{table_full_landsat_all}}
\resizebox{\linewidth}{!}  
{
\begin{tabular}{c|ccc|ccc|ccc|ccc}
\hline
& \multicolumn{3}{c|}{Full-Landsat-800}                        & \multicolumn{3}{c|}{Full-Landsat-1200}                       
& \multicolumn{3}{c|}{Full-Landsat-1600}    
& \multicolumn{3}{c}{Full-Landsat-2000} 
\\ \cline{2-13} 
\multirow{-2}{*}{Method} 
& $D_{\lambda}$ $\downarrow$        & $D_{S}$ $\downarrow$   & QNR $\uparrow$      
& $D_{\lambda}$ $\downarrow$        & $D_{S}$ $\downarrow$   & QNR $\uparrow$    
& $D_{\lambda}$ $\downarrow$        & $D_{S}$ $\downarrow$   & QNR $\uparrow$    
& $D_{\lambda}$ $\downarrow$        & $D_{S}$ $\downarrow$   & QNR $\uparrow$ \\ \hline
GS & 0.0519 & 0.1145 & 0.8405 & 0.0589 & 0.1155 & 0.8334 & 0.0567 & 0.1076 & 0.8425 & 0.0598 & 0.1064 & 0.8406 \\
IHS & 0.0862 & 0.1235 & 0.8012 & 0.0972 & 0.1286 & 0.7883 & 0.0982 & 0.1205 & 0.7945 & 0.1028 & 0.1205 & 0.7905 \\
GFPCA & 0.0731 & 0.0577 & 0.8737 & 0.0623 & 0.0543 & 0.8870 & 0.0627 & 0.0525 & 0.8883 & 0.0615 & 0.0510 & 0.8910 \\ \hline
MSDCNN & 0.0523 & 0.0378 & 0.9135 & 0.0476 & 0.0316 & 0.9234 & 0.0485 & 0.0307 & 0.9234 & 0.0468 & 0.0303 & 0.9254 \\
SFINet & 0.0493 & 0.0405 & 0.9137 & 0.0410 & 0.0331 & 0.9278 & 0.0404 & 0.0314 & 0.9299 & 0.0392 & 0.0317 & 0.9307 \\
MSDDN & 0.0402 & 0.0246 & 0.9372 & 0.0311 & \underline{0.0224} & 0.9477 & 0.0317 & 0.0229 & 0.9466 & 0.0312 & 0.0246 & 0.9454 \\
PanFlowNet & 0.0570 & 0.0388 & 0.9077 & 0.0449 & 0.0379 & 0.9194 & 0.0458 & 0.0385 & 0.9178 & 0.0449 & 0.0384 & 0.9187 \\
HFIN & 0.0287 & \textbf{0.0200} & \underline{0.9523} & \textbf{0.0228} & \textbf{0.0201} & 0.9543 & \underline{0.0229} & \textbf{0.0188} & \textbf{0.9590} & \underline{0.0217} & \underline{0.0198} & \underline{0.9591} \\
Pan-mamba & 0.0295 & 0.0525 & 0.9196 & 0.0297 & 0.0675 & \underline{0.9049} & 0.0312 & 0.0816 & 0.8898 & 0.0330 & 0.0928 & 0.8773 \\
ARConv & \underline{0.0278} & \underline{0.0231} & 0.9501 & 0.0260 & 0.0200 & 0.9548 & 0.0251 & \underline{0.0191} & 0.9566 & 0.0239 & \textbf{0.0183} & 0.9585 \\
\rowcolor{gray!20}
Ours & \textbf{0.0272} & 0.0271 & \textbf{0.9537} & \underline{0.0240} & 0.0225 & \textbf{0.9563} & \textbf{0.0217} & 0.0209 & \underline{0.9580} & \textbf{0.0206} & 0.0206 & \textbf{0.9594} \\
\hline
\end{tabular}
}
\end{table*}

\begin{table*}[htpb]
\centering
\small  
\renewcommand{\arraystretch}{1.1}
\renewcommand{\tabcolsep}{3pt}  
\caption{Multi-resolution results on the Full-Skysat datasets, with numbers indicating the size of the PAN image. The best are highlighted in \textbf {bold}, and the second are \underline {underlined}.\label{table_full_skysat_all}}
\resizebox{\linewidth}{!}  
{
\begin{tabular}{c|ccc|ccc|ccc|ccc}
\hline
& \multicolumn{3}{c|}{Full-Skysat-800}                        & \multicolumn{3}{c|}{Full-Skysat-1200}                       
& \multicolumn{3}{c|}{Full-Skysat-1600}    
& \multicolumn{3}{c}{Full-Skysat-2000} 
\\ \cline{2-13} 
\multirow{-2}{*}{Method} 
& $D_{\lambda}$ $\downarrow$        & $D_{S}$ $\downarrow$   & QNR $\uparrow$      
& $D_{\lambda}$ $\downarrow$        & $D_{S}$ $\downarrow$   & QNR $\uparrow$    
& $D_{\lambda}$ $\downarrow$        & $D_{S}$ $\downarrow$   & QNR $\uparrow$    
& $D_{\lambda}$ $\downarrow$        & $D_{S}$ $\downarrow$   & QNR $\uparrow$ \\ \hline
GS & 0.0533 & 0.1033 & 0.8499 & 0.0657 & 0.1163 & 0.8273 & 0.0653 & 0.1129 & 0.8304 & 0.0649 & 0.1123 & 0.8311 \\
IHS & 0.0937 & 0.1282 & 0.7916 & 0.0973 & 0.1343 & 0.7834 & 0.0993 & 0.1333 & 0.7822 & 0.0994 & 0.1314 & 0.7836 \\
GFPCA & 0.0427 & 0.0633 & 0.8974 & 0.0445 & 0.0633 & 0.8956 & 0.0428 & 0.0620 & 0.8983 & 0.0432 & 0.0610 & 0.8989 \\ \hline
MSDCNN & 0.0449 & \textbf{0.0262} & \underline{0.9344} & 0.0407 & \underline{0.0282} & \underline{0.9422} & 0.0422 & \textbf{0.0187} & \underline{0.9391} & 0.0398 & \textbf{0.0155} & \textbf{0.9455} \\
SFINet & 0.1357 & 0.0331 & 0.8373 & 0.1304 & 0.0343 & 0.8412 & 0.1363 & 0.0333 & 0.8362 & 0.1361 & 0.0327 & 0.8368 \\
MSDDN & \underline{0.0424} & 0.0386 & 0.9217 & \underline{0.0388} & 0.0382 & 0.9250 & 0.0426 & 0.0355 & 0.9294 & 0.0408 & 0.0327 & 0.9293 \\
PanFlowNet & 0.0883 & 0.0376 & 0.8787 & 0.0761 & 0.0339 & 0.8935 & 0.0738 & 0.0312 & 0.8981 & 0.0726 & 0.0301 & 0.9003 \\
HFIN & 0.0623 & 0.0634 & 0.8796 & 0.0583 & 0.0573 & 0.8888 & 0.0584 & 0.0588 & 0.8872 & 0.0572 & 0.0565 & 0.8904 \\
Pan-mamba & 0.0579 & 0.0383 & 0.9071 & 0.0546 & 0.0330 & 0.9152 & 0.0515 & 0.0284 & 0.9224 & 0.0491 & \underline{0.0248} & 0.9279 \\
ARConv & \textbf{0.0405} & 0.0416 & 0.9201 & \textbf{0.0381} & 0.0402 & 0.9237 & \textbf{0.0380} & 0.0415 & 0.9225 & \underline{0.0380} & 0.0404 & 0.9236 \\
\rowcolor{gray!20}
Ours & 0.0451 & \underline{0.0292} & \textbf{0.9365} & 0.0423 & \textbf{0.0248} & \textbf{0.9425} & \underline{0.0407} & \underline{0.0269} & \textbf{0.9426} & \textbf{0.0339} & 0.0268 & \underline{0.9416} \\
\hline
\end{tabular}
}
\end{table*}
\end{document}